%% file: arxiv.tex
\documentclass{bmvc2k}

\input{sections/import_package.tex}

\input{sections/0_header.tex}

\def\eg{\emph{e.g}\bmvaOneDot}

\begin{document}

\maketitle

\input{sections/abstract.tex}

\input{sections/1_intro.tex}

\input{sections/2_related_work.tex}

\input{sections/3_method.tex}

\input{sections/4_experiment.tex}

\input{sections/5_conclusion.tex}

\section*{Acknowledgements}
This work was supported by the National Research Foundation of Korea (NRF) grant (No. 2021R1A2C2012195),
Institute of Information \& Communications Technology
Planning \& Evaluation (IITP) grant (2021–0–00537, Visual
Common Sense Through Self-supervised Learning for
Restoration of Invisible Parts in Images), and IITP grant
(2020–0–01336, Artificial Intelligence Graduate School Program, UNIST), all funded by the Korea government (MSIT).
We thank Yunpyo An and Jihun Lee for their assistance in conducting a qualitative analysis of the results.

\clearpage
\section*{Appendix}
\input{sections/supplementary.tex}

\bibliography{egbib}
\end{document}

%% file: sections/import_package.tex
\usepackage{graphicx}
\usepackage{amsmath}
\usepackage{amssymb}
\usepackage{algorithm}
\usepackage{algpseudocode}

\usepackage{tikz}
\usetikzlibrary{fit,backgrounds}
\usetikzlibrary{quotes,arrows.meta}
\usetikzlibrary{positioning}
\usetikzlibrary{3d}
\usepackage{adjustbox}
\usepackage{caption}
\usepackage{multirow}
\usepackage{tabularx}

%% file: sections/0_header.tex
\title{BoIR: Box-Supervised Instance Representation for Multi-Person Pose Estimation}

\addauthor{Uyoung Jeong}{jeong.uyoung@unist.ac.kr}{1}
\addauthor{Seungryul Baek}{srbaek@unist.ac.kr}{1}
\addauthor{Hyung Jin Chang}{h.j.chang@bham.ac.uk}{2}
\addauthor{Kwang In Kim}{kimkin@postech.ac.kr}{3}

\addinstitution{
Ulsan National Institute of Science and Technology\\
 Ulsan, Republic of Korea
}
\addinstitution{
University of Birmingham\\
Birmingham, United Kingdom
}
\addinstitution{
Pohang University of Science and Technology\\
 Pohang, Republic of Korea
}

\runninghead{Jeong, Baek, Chang, Kim}{BoIR: Box-Supervised Instance Representation}

%% file: sections/abstract.tex
\begin{abstract}
Single-stage multi-person human pose estimation (MPPE) methods have shown great performance improvements, but existing methods fail to disentangle features by individual instances under crowded scenes. In this paper, we propose a bounding box-level instance representation learning called BoIR, which simultaneously solves instance detection, instance disentanglement, and instance-keypoint association problems. Our new instance embedding loss provides a learning signal on the entire area of the image with bounding box annotations, achieving globally consistent and disentangled instance representation. Our method exploits multi-task learning of bottom-up keypoint estimation, bounding box regression, and contrastive instance embedding learning, without additional computational cost during inference. BoIR is effective for crowded scenes, outperforming state-of-the-art on COCO val (0.8 AP), COCO test-dev (0.5 AP), CrowdPose (4.9 AP), and OCHuman (3.5 AP).
Code will be available at \url{https://github.com/uyoung-jeong/BoIR}
\end{abstract}

%% file: sections/1_intro.tex
\section{Introduction}
Multi-person human pose estimation(MPPE) localizes 2D keypoint locations of multiple human instances from an image.
It is useful not only for 3D pose estimation and activity recognition~\cite{Zhang_2020_CVPR}, but also for human-robot interaction~\cite{8643076}, autonomous driving~\cite{Zheng_2022_CVPR}, augmented/virtual reality and surveillance applications. In wild scenarios, where severe inter-person occlusion and background clutter frequently occur, the capability of multi-person pose estimation becomes even more crucial.

\input{figures/6_bml_vis}
Recent advances in single-stage MPPE methods~\cite{Mao_2021_CVPR, Wang_2022_CVPR,yang2023explicit} have shown significant performance improvements. Compared to top-down methods~\cite{Khirodkar_2021_ICCV,xiao2018simple,CrowdPose_CVPR19}, they do not require off-the-shelf person detectors and therefore robust to detection errors. Unlike bottom-up methods~\cite{Cheng_2020_CVPR, Luo_2021_CVPR, Geng_2021_CVPR, Xue_2022_CVPR, ce_2022_eccv}, they solve instance-keypoint association problems by explicitly detecting instances, usually using instance center locations.

While single-stage methods showed promising results, they still suffer from instance-keypoint association under heavy inter-person occlusion, which often results in noisy predictions. We summarize the main reasons in two aspects. 
First, existing representation-based methods lack multi-task supervision to learn diverse aspects of instance representation. Even if they add multiple tasks, it would incur computational overhead during inference.
Second, previous works have spatially sparse supervision. Many works apply losses only on ground-truth keypoint locations, which is too sparse for the model to holistically learn the entire image region, leading to noisy and globally inconsistent results, as illustrated in Fig.~\ref{fig:bml_illustration}. Although heatmap-based approaches apply Gaussian kernel to generate ground-truth keypoint heatmaps, it is still more sparse than conventional segmentation level supervision.

In this paper, we focus on an effective instance representation learning method which can provide both rich spatial and multi-task supervision.
First, we reformulate the existing MPPE pipeline to apply embedding loss on a separate embedding branch, which can effectively map nonlinear features of instances while the primary task branch's performance is not degraded.
Then, we design a new contrastive learning scheme, termed Bbox Mask Loss, using bounding box(bbox) supervision. It contrasts instance embeddings on both inside and outside of the ground-truth boxes, which provides learning signals on the entire image region.
Combining with box regression and bottom-up keypoint heatmap regression as auxiliary tasks, we apply multi-task learning scheme to learn effective instance representation for multiple keypoint estimation.

In summary, we introduce a novel method for instance representation learning at the box level, named BoIR. BoIR adeptly addresses the challenges of instance disentanglement and instance detection simultaneously, without incurring any additional computational costs during inference. These are achieved through the following key contributions:
\begin{itemize}
    \item Bbox Mask Loss effectively disentangles features by instances in the embedding space using a new embedding loss with spatially rich bounding box level supervision.
    \item Auxiliary task heads enrich instance representation by sharing multiple aspects of the instance, while no additional computational cost is induced during inference.
\end{itemize}
BoIR excels at challenging crowded scenes, surpassing comparative methods by 0.5 AP on COCO \texttt{test-dev}, 4.9 AP on CrowdPose \texttt{test}, and 3.5 AP on OCHuman \texttt{test}.


%% file: figures/6_bml_vis.tex
\begin{figure}[!ht]
\begin{center}
\begin{tabular}{c@{\hspace{0.3em}}c@{\hspace{0.3em}}c@{\hspace{0.3em}}c@{\hspace{0.3em}}c@{\hspace{0.3em}}c}
\raisebox{-0.5\height}{\includegraphics[width=0.22\linewidth]{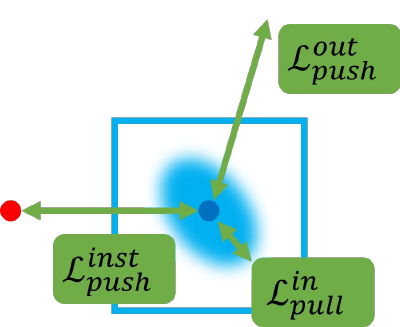}} &
\raisebox{-0.5\height}{\includegraphics[width=0.14\linewidth]{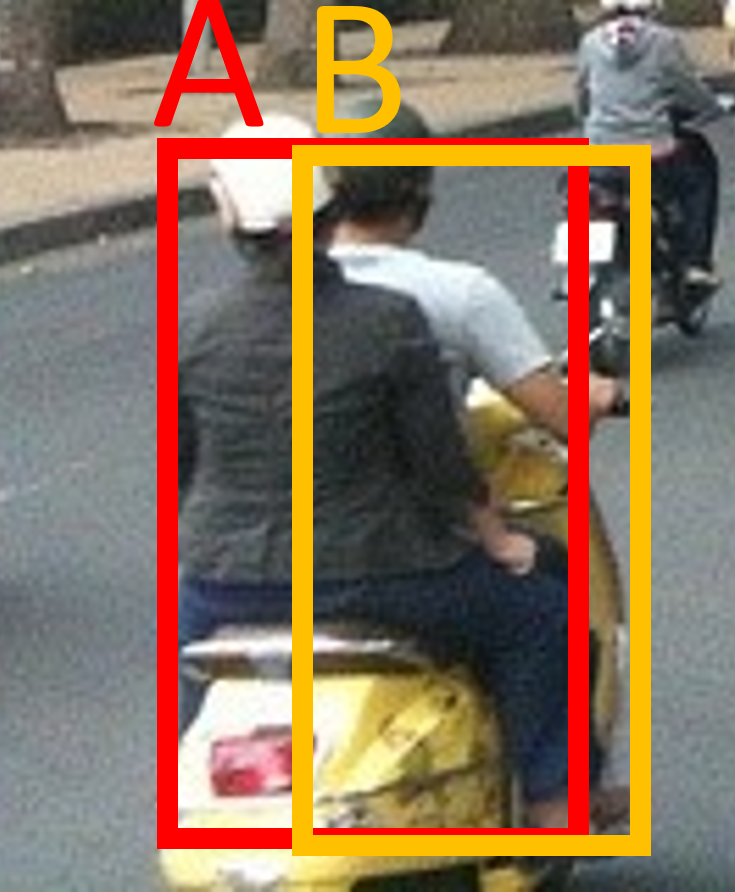}} &
\raisebox{-0.5\height}{\includegraphics[width=0.14\linewidth]{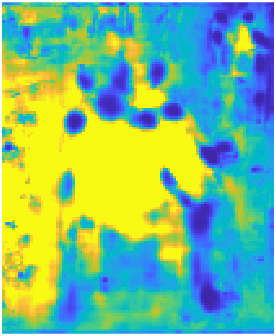}} &
\raisebox{-0.5\height}{\includegraphics[width=0.14\linewidth]{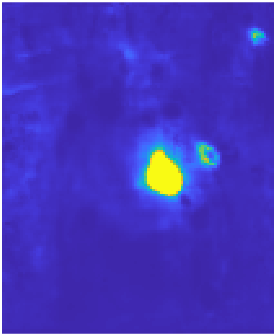}} &  
\raisebox{-0.5\height}{\includegraphics[width=0.14\linewidth]{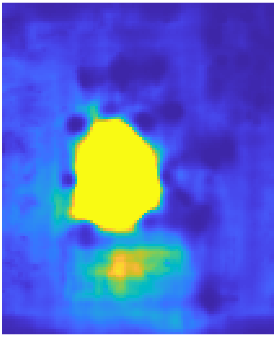}} &
\raisebox{-0.5\height}{\includegraphics[width=0.14\linewidth]{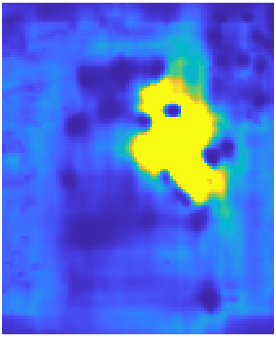}} \\
(a) & (b) & (c) & (d) & (e) & (f)
\end{tabular}
\end{center}
\caption{(a): Bbox Mask Loss framework. Blue dot is a query box(blue colour) center, while red dot is another box center. $\mathcal{L}_{pull}^{in}$ pulls instance center and soft-masked mean embeddings inside the box, $\mathcal{L}_{push}^{out}$ pushes pairwise instance-background embeddings, and $\mathcal{L}_{push}^{inst}$ pushes pairwise instance embeddings.
(b)-(f): Visualization of feature similarities from the center features of bounding boxes in (b). (c) and (d) are CID feature similarities from A and B centers, respectively, while (e) and (f) are BoIR feature similarities.}
\label{fig:bml_illustration}
\end{figure}

%% file: sections/2_related_work.tex
\section{Related Works}
\noindent \textbf{2D Multi-person human pose estimation (MPPE).}
2D MPPE methods can be roughly classified by instance handling approaches.
Top-down methods use detectors~\cite{he2017mask,redmon2018yolov3,ren2015faster} to get person boxes and use cropped images as input.
Bottom-up methods first detect keypoints and group them into instances.
Single-stage methods, on the other hand, detect instances first and then regress instance-wise keypoints.
Single-stage methods eliminate the need to crop an image into multiple instance-wise images, and avoid the need for keypoint grouping.

SimpleBaseline~\cite{xiao2018simple} and HRNet~\cite{Sun_2019_CVPR} are top-down methods, and are generally used as backbone networks in various works.
MIPNet~\cite{Khirodkar_2021_ICCV} is one of the recent top-down approaches that considers multiple instances within a box by modulating the channel dimensions to regress individual keypoints.

OpenPose~\cite{cao2017realtime}, PersonLab~\cite{papandreou2018personlab}, and PifPaf~\cite{Kreiss_2019_CVPR} share a similar idea of estimating a vector field that associates keypoints with instances.
HigherHRNet~\cite{Cheng_2020_CVPR} and its subsequent works~\cite{Luo_2021_CVPR, Geng_2021_CVPR, Xue_2022_CVPR, ce_2022_eccv} are another class of bottom-up methods using Associative Embedding~\cite{NIPS2017_8edd7215}. From the pixel-wise one-dimensional embedding, they assign the detected keypoints to respective instances using off-the-shelf grouping algorithm~\cite{kuhn1955hungarian}. These methods tend to lack the capability of instance detection since their training losses are mainly targeted for keypoint estimation.

There are several single-stage methods based on Transformers~\cite{vaswani2017attention}. PETR~\cite{Shi_2022_CVPR} avoids using Hungarian algorithm for instance grouping by randomly initializing query embeddings to regress keypoints. In contrast, ED-Pose~\cite{yang2023explicit} extracts query embeddings via a human detection decoder, but it requires substantial computational cost due to the massive amount of learnable parameters, which is critical for real-time pose estimation. QueryPose~\cite{xiao2022querypose} similarly performs box and keypoint regression via query embeddings and Transformers-based decoders, and its performance on CrowdPose \texttt{test} is inferior to CID by 0.2 AP with the same HRNet-W48 backbone.

FCPose~\cite{Mao_2021_CVPR} and CID~\cite{Wang_2022_CVPR} are single-stage methods using an instance center map.
FCPose generates instance proposals from a single-stage person detector and employs instance-wise dynamic convolution on global features.
Similarly, CID estimates instance center map to detect instances, and performs channel and spatial attention between sampled feature and global features, but it does not perform box regression.
CID directly applies contrastive loss on the backbone network's output feature, which does not effectively disentangle features by instances, as discussed in SimCLR~\cite{chen2020simple}. Additionally, CID's contrastive loss is spatially sparse since it is applied solely on instance center locations. Instead, we introduce a separate embedding branch that enhances learning keypoint features, providing richer spatial and multi-task guidance.
KAPAO~\cite{kapao_2022_eccv} is another method that reformulates keypoint regression task as an object detection task, jointly detecting persons and keypoints.

\vspace{1mm}
\noindent \textbf{Representation learning with distance metrics.}
Deep metric learning aims to learn a distance metric in the embedding space for better representation, generally composed with a pull term for closing the distance among positive samples, and a push term for differentiating between different classes. Push loss term is crucial for effective representation learning, so many works are devoted to proposing various negative sampling strategies. Contrastive loss~\cite{hadsell2006dimensionality}, triplet loss~\cite{schroff2015facenet}, N-pair loss~\cite{NIPS2016_6b180037} and InfoNCE loss~\cite{oord2018representation} are some of the approaches. SimCLR~\cite{chen2020simple}, MoCo~\cite{he2020momentum}, and CLIP~\cite{radford2021learning} are representative works using variants of InfoNCE loss. All of these methods use cosine similarity as a similarity metric.

%% file: sections/3_method.tex
\input{figures/1_pipeline.tex}
\vspace{-3mm}
\section{Method}
\subsection{Framework overview}
Our framework comprises two main parts: auxiliary task branch and instance keypoint branch. Given an input image, backbone network outputs a feature $f\in\mathbb{R}^{C,H,W}$, where $H$ is height and $W$ is width. Task-specific heads produce instance center heatmaps $c\in\mathbb{R}^{1,H,W}$, box predictions $b\in\mathbb{R}^{4,H,W}$, bottom-up keypoint heatmaps $k^{bu}\in\mathbb{R}^{K,H,W}$ and instance embedding map $e\in\mathbb{R}^{D,H,W}$. 
During training, after detecting instances from the center map, instance features $f^{p}$ are sampled from the backbone feature at the ground-truth center coordinates. $f^p$ are used as conditions for regressing instance-wise keypoints $k$ in the instance keypoint head, as proposed in~\cite{Wang_2022_CVPR}. In case of embedding branch, we sample instance embeddings $p$ from $e$.
During inference, $f^{p}$ are sampled from predicted instance centers. We made several enhancements to the instance keypoint head, including Layer Normalization and Instance Normalization for stable learning, as illustrated in Fig.~\ref{fig:framework_overview}. Please note that $b, k^{bu}, e$ are not estimated during inference.

\subsection{Bbox Mask Loss}
\label{section:bbox_mask_loss}
Existing instance representation learning methods such as Associative Embedding(AE) and CID's contrastive loss failed to handle multiple people in several aspects, often leading to noisy results. Firstly, they only compare instance embeddings with ground-truth(GT) instance locations, making it difficult to generate a push loss term when only one GT instance is present in an image. Secondly, there are unlabeled instances in training datasets, and existing works typically ignore these unlabeled instances, which induces additional noise during inference. Thirdly, the number of human instances per image in training datasets is insufficient for effective instance representation learning. For example, COCO \texttt{train} set has an average of 2.6 people per image, excluding labels with \texttt{iscrowd}=1. Similarly, CrowdPose \texttt{trainval} set has 4.2 people per image.

To alleviate aforementioned challenges, inspired by a weakly supervised instance segmentation method~\cite{wolny2022sparse}, we introduce spatially rich supervision via box annotations, termed Bbox Mask Loss. It disambiguates each instance embedding from outside of the box region, which can handle arbitrary unlabeled instances and background clutter. It applies soft masking on the inside of the box based on embedding similarity, which is effective for feature disentanglement under heavy cross-instance occlusions. Moreover, it can produce push loss term even when only a single GT instance is available in an image, serving as a simple but effective negative sampling method.

Bbox Mask Loss incorporates multitude of push and pull loss terms, including in-box pull $\mathcal{L}^{in}_{pull}$, out-box push $\mathcal{L}^{out}_{push}$, and cross-instance push $\mathcal{L}^{inst}_{push}$.
First, given a GT instance and corresponding box with height $h$ and width $w$, we compute pixel-wise embedding similarity between embedding map and the instance embedding as follows: 
\begin{align}
s_{i}^{(x,y)} = \psi(d(e^{(x,y)},p_i)), \quad (x,y)\in \mathcal{B}_{i},
\label{eq:1}
\end{align}
where $d$ is a distance metric, and $\psi$ is an inversion operator to convert the distance to similarity with [0,1] output range. From ablative experiment, as reported in Table~\ref{table:ablation_main}, we find that L2 distance for $d$ and Gaussian kernel for $\psi$ outperforms cosine distance and cosine similarity. $\mathcal{B}_{i}$ is a set of coordinates inside the box $b_{i}$, where $i=1,2,...,N$.
As a pulling term inside the box, we want the model to produce similar embeddings on the foreground region of the same person. To realize the objective, we compare the instance center embedding with the mean instance embedding $\bar{p}_{i}$, as defined below:
\begin{align}
\mathcal{L}^{in}_{pull} = \frac{1}{N}\sum_{i=1}^{N}{d(p_i, \bar{p}_i)}, \quad \text{where} \quad \bar{p}_i = \frac{\sum_{(x,y)\in\mathcal{B}_{i}}e^{(x,y)}s^{(x,y)}_{i}}{\sum_{(x,y)\in\mathcal{B}_{i}}s^{(x,y)}_{i}}.
\label{eq:2}
\end{align}
To decouple the instance embedding from the background, we define the out-box push loss using out-box mean embedding $\bar{p}^{c}_{i}$: 
\begin{align}
\mathcal{L}^{out}_{push} = \frac{1}{N}\sum_{i=1}^{N}\psi(d(p_i, \bar{p}^{c}_{i})), \quad \text{where} \quad\bar{p}^{c}_{i}=\frac{\sum_{(x,y)\in\mathcal{B}^{c}_{i}}e_{(x,y)}}{|\mathcal{B}^{c}_{i}|}.
\label{eq:3}
\end{align}
Note that $\mathcal{B}^{c}_{i}$ is a set of coordinates outside the $i$th box, and $\bar{p}^{c}_{i}$ is a mean embedding of the background except the $i$th box region. Lastly, cross-instance push term compares instance embeddings retrieved from ground-truths, which is the same as the existing losses.
\begin{align}
\mathcal{L}^{inst}_{push} = \frac{1}{(N(N-1)/2)}\sum_{i=1}^{N}\sum_{j>i}^{N}\psi(d(p_i,p_{j})).
\label{eq:4}
\end{align}

\subsection{Auxiliary tasks}
In order to encourage the features to have richer and more disentangled information for MPPE, we incorporate multiple auxiliary tasks and instance representation learning in parallel. Our multi-task branch consists of shared layers and four separate regression heads, consisting of instance embedding, bottom-up keypoint, box, and instance center.

We concurrently reduce dimensionality of the backbone feature and incorporate multi-resolution shared feature representation based on ASPPv2~\cite{aspp_2017_arxiv}. It resolves the problem of regressing globally consistent instance features. Original ASPPv2 module incurs heavy computational cost when fusing multiple resolution features. We alleviate this by further squeezing the output channel size of each multi-resolution feature to 128, and then apply a fusion layer to obtain a final feature with 256 channel size. This design reduces the number of trainable parameters of ASPP by 50\%. This shared bottleneck module design helps to prevent auxiliary tasks from dominating over the primary task, by restricting the amount of information flow to the auxiliary tasks. 

Each regression head comprises with one residual block and one output convolution layer for sufficient capability of learning nonlinear feature transformation. In case of box regression, we adopt anchor free method~\cite{li2022yolov6} for efficient training. For clarity, we do not use the bbox head outputs during inference, and the box head serves as an efficient and informative auxiliary task head.

\input{tables/coco_testdev}
\subsection{Training losses}
We employ five loss functions: instance-wise keypoint heatmap loss $\mathcal{L}_{kpt}$, center heatmap loss $\mathcal{L}_{center}$, bottom-up keypoint heatmap loss $\mathcal{L}_{buk}$, bbox loss $\mathcal{L}_{bbox}$, and embedding loss $\mathcal{L}_{emb}$.
\begin{align}
\mathcal{L} = \mathcal{L}_{kpt} + \mathcal{L}_{center} + \mathcal{L}_{buk} + \mathcal{L}_{bbox} + \mathcal{L}_{emb}.
\end{align}
Focal loss~\cite{Law_2018_ECCV,objects_as_points} is used for $\mathcal{L}_{kpt}, \mathcal{L}_{center}$ and $\mathcal{L}_{buk}$, while CIoU loss~\cite{zheng2020distance} is used for $\mathcal{L}_{bbox}$. 
For embedding loss, we use three loss terms as defined in Equation~\ref{eq:2},\ref{eq:3},\ref{eq:4}. We use AE loss for calculating respective terms:
\begin{align}
\mathcal{L}_{emb} = \mathcal{L}^{in}_{pull} + \mathcal{L}^{out}_{push} + \mathcal{L}^{inst}_{push}.
\end{align}

%% file: figures/1_pipeline.tex
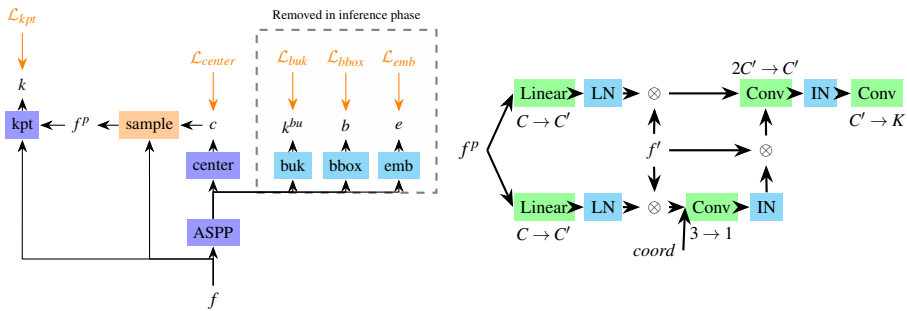
\begin{figure}[!ht]
\begin{center}
\begin{tabular}{c@{\hspace{0.3em}}c}
\raisebox{-0.5\height}{
\begin{tikzpicture}[thick,scale=0.7, every node/.style={transform shape}]
\tikzstyle{connection}=[ultra thick, every node/.style={sloped,allow upside down, minimum size=1em},draw=\edgecolor,opacity=0.7]
\node[canvas is xy plane at z=0] (f) at (0,0,0) {$f$};

\node[shift={(0,1.0,0)},fill=blue!40,minimum width=1,minimum height=1.4em,font=\small] at (f.north) (neck) {ASPP};

\node[shift={(0,1.0,0)},fill=blue!40,minimum width=1,minimum height=1.4em,font=\small] at (neck.north) (center_head) {center};

\node[shift={(1.5,1.0,0)},fill=cyan!40,minimum width=1,minimum height=1.4em,font=\small] at (neck.north) (buk_head) {buk};

\node[shift={(2.5,1.0,0)},fill=cyan!40,minimum width=1,minimum height=1.4em,font=\small] at (neck.north) (bbox_head) {bbox};

\node[shift={(3.5,1.0,0)},fill=cyan!40,minimum width=1,minimum height=1.4em,font=\small] at (neck.north) (emb_head) {emb};

\node[shift={(0,0.5,0)},minimum width=1,minimum height=1.4em,font=\small] at (center_head.north) (center) {$c$};

\node[shift={(0,0.5,0)},minimum width=1,minimum height=1.4em,font=\small] at (buk_head.north) (buk) {$k^{bu}$};

\node[shift={(0,0.5,0)},minimum width=1,minimum height=1.4em,font=\small] at (bbox_head.north) (bbox) {$b$};

\node[shift={(0,0.5,0)},minimum width=1,minimum height=1.4em,font=\small] at (emb_head.north) (emb) {$e$};

\node[shift={(-1.0,0,0)},fill=orange!40,minimum width=1,minimum height=1.4em,font=\small] at (center.west) (sample) {sample};

\node[shift={(-0.71,0,0)},minimum width=1,minimum height=1.4em,font=\small] at (sample.west) (fp) {$f^p$};

\node[shift={(-0.805,0,0)},fill=blue!40,minimum width=1,minimum height=1.4em,font=\small] at (fp.west) (kpt_head) {kpt};

\node[shift={(0,0.5,0)},minimum width=1,minimum height=1.4em,font=\small] at (kpt_head.north) (kpt) {$k$};

\node[shift={(0,1.0,0)},minimum width=1,minimum height=1.4em,font=\small, text=orange] at (kpt.north) (kpt_loss) {$\mathcal{L}_{kpt}$};

\node[shift={(0,1.0,0)},minimum width=1,minimum height=1.4em,font=\small, text=orange] at (center.north) (center_loss) {$\mathcal{L}_{center}$};

\node[shift={(0,1.0,0)},minimum width=1,minimum height=1.4em,font=\small, text=orange] at (buk.north) (buk_loss) {$\mathcal{L}_{buk}$};

\node[shift={(0,1.0,0)},minimum width=1,minimum height=1.4em,font=\small, text=orange] at (bbox.north) (bbox_loss) {$\mathcal{L}_{bbox}$};

\node[shift={(0,1.0,0)},minimum width=1,minimum height=1.4em,font=\small, text=orange] at (emb.north) (emb_loss) {$\mathcal{L}_{emb}$};

\begin{pgfonlayer}{background}
\node[draw=gray,dashed,fill=gray!1,thick, inner sep=5pt,fit=(buk_head) (bbox_head) (emb_head) (buk) (bbox) (emb) (buk_loss) (bbox_loss) (emb_loss)] (trim) {};
\node[shift={(0,0.25,0)},minimum width=1,font=\scriptsize] at (trim.north) (dashed_box_caption) {Removed in inference phase};

\end{pgfonlayer}
\draw [-Stealth]  (f.north)    -- (neck.south);
\draw [-Stealth]  (neck.north)    -- (center_head.south);
\draw [-Stealth]  (neck.north) -- ++(0,0.5,0)  -- ++(1.5,0,0)  -- (buk_head.south);
\draw [-Stealth]  (neck.north) -- ++(0,0.5,0) -- ++(2.5,0,0)  -- (bbox_head.south);
\draw [-Stealth]  (neck.north) -- ++(0,0.5,0) -- ++(3.5,0,0)  -- (emb_head.south);

\draw [-Stealth]  (center_head.north)    -- (center.south);
\draw [-Stealth]  (buk_head.north)    -- (buk.south);
\draw [-Stealth]  (bbox_head.north)    -- (bbox.south);
\draw [-Stealth]  (emb_head.north)    -- (emb.south);

\draw [-Stealth]  (f.north) -- ++(0,0.5,0) -- ++(-1.2,0,0)  -- (sample.south);
\draw [-Stealth]  (f.north) -- ++(0,0.5,0) -- ++(-3.6,0,0)  -- (kpt_head.south);
\draw [-Stealth]  (center.west)    -- (sample.east);
\draw [-Stealth]  (sample.west)    -- (fp.east);
\draw [-Stealth]  (fp.west)    -- (kpt_head.east);
\draw [-Stealth]  (kpt_head.north)    -- (kpt.south);

\draw [-Stealth, fill=orange,draw=orange]  (kpt_loss.south)    -- (kpt.north);
\draw [-Stealth, fill=orange,draw=orange]  (center_loss.south)    -- (center.north);
\draw [-Stealth, fill=orange,draw=orange]  (buk_loss.south)    -- (buk.north);
\draw [-Stealth, fill=orange,draw=orange]  (bbox_loss.south)    -- (bbox.north);
\draw [-Stealth, fill=orange,draw=orange]  (emb_loss.south)    -- (emb.north);
\end{tikzpicture}} &
\raisebox{-0.5\height}{
\input{figures/5_ikh_comp}
}
\end{tabular}
\end{center}
\caption{Left: Overview of our framework. Instance keypoint (kpt) head and center head are primary regression heads. bottom-up keypoint (buk) head, bounding box (bbox) head and embedding (emb) head are auxiliary task regressors which are not used during inference. Right: Layer composition of instance keypoint head. 'Linear': linear layer, 'Conv': convolution layer, 'LN': Layer Normalization, 'IN': Instance Normalization, '$\otimes$': Hadamard product. '$coord$': relative coordinates of the heatmap pixel indices. $f'\in\mathbb{R}^{C'\times H\times W}$: projection of $f$ by single convolution layer.}
\label{fig:framework_overview}
\end{figure}

%% file: figures/5_ikh_comp.tex
\begin{tikzpicture}[thick,scale=0.75, every node/.style={transform shape}]
\tikzstyle{connection}=[ultra thick,every node/.style={sloped,allow upside down, minimum size=1em},draw=\edgecolor,opacity=0.7]
\node[canvas is xy plane at z=0] (fp) at (0,0,0) {$f^p$};

\node[shift={(1.0,1,0)},fill=green!40,minimum width=1,minimum height=1.4em,font=\small] at (fp.east) (c_linear) {Linear};
\node[shift={(0,-0.2,0)},minimum width=1,minimum height=1.4em,font=\small] at (c_linear.south) (c_linear_dim) {$C \rightarrow C'$};

\node[shift={(0.5,0,0)},fill=cyan!40,minimum width=1,minimum height=1.4em,font=\small] at (c_linear.east) (c_ln) {LN};

\node[shift={(0.9,0,0)},minimum width=1,minimum height=1.4em,font=\small] at (c_ln) (c_hadamard) {$\otimes$};

\node[shift={(2.935,0,0)},minimum width=1,minimum height=1.4em,font=\small] at (fp.east) (f) {$f'$};

\node[shift={(1.0,-1.0,0)},fill=green!40,minimum width=1,minimum height=1.4em,font=\small] at (fp.east) (s_linear) {Linear};
\node[shift={(0,-0.2,0)},minimum width=1,minimum height=1.4em,font=\small] at (s_linear.south) (s_linear_dim) {$C \rightarrow C'$};

\node[shift={(0.5,0,0)},fill=cyan!40,minimum width=1,minimum height=1.4em,font=\small] at (s_linear.east) (s_ln) {LN};

\node[shift={(0.9,0,0)},minimum width=1,minimum height=1.4em,font=\small] at (s_ln) (s_hadamard) {$\otimes$};

\node[shift={(0,-0.5,0)},minimum width=1,minimum height=1.4em,font=\small] at (s_hadamard.south) (coord) {$coord$};

\node[shift={(0.75,0,0)},fill=green!40,minimum width=1,minimum height=1.4em,font=\small] at (s_hadamard.east) (s_conv) {Conv};
\node[shift={(0,-0.2,0)},minimum width=1,minimum height=1.4em,font=\small] at (s_conv.south) (s_conv_dim) {$3 \rightarrow 1$};

\node[shift={(0.5,0,0)},fill=cyan!40,minimum width=1,minimum height=1.4em,font=\small] at (s_conv.east) (s_in) {IN};

\node[shift={(-0.01,1.0,0)},minimum width=1,minimum height=1.4em,font=\small] at (s_in) (s_hadamard2) {$\otimes$};

\node[shift={(1.7,0,0)},fill=green!40,minimum width=1,minimum height=1.4em,font=\small] at (c_hadamard.east) (fusion_conv) {Conv};
\node[shift={(0,0.2,0)},minimum width=1,minimum height=1.4em,font=\small] at (fusion_conv.north) (fusion_conv_dim) {$2C' \rightarrow C'$};

\node[shift={(0.5,0,0)},fill=cyan!40,minimum width=1,minimum height=1.4em,font=\small] at (fusion_conv.east) (fusion_in) {IN};

\node[shift={(0.7,0,0)},fill=green!40,minimum width=1,minimum height=1.4em,font=\small] at (fusion_in.east) (last_conv) {Conv};
\node[shift={(0,-0.2,0)},minimum width=1,minimum height=1.4em,font=\small] at (last_conv.south) (last_conv_dim) {$C' \rightarrow K$};

\draw [-Stealth]  (fp.east) -- (c_linear.west);
\draw [-Stealth]  (c_linear.east) -- (c_ln.west);
\draw [-Stealth]  (c_ln.east) -- (c_hadamard.west);
\draw [-Stealth]  (f.north) -- (c_hadamard.south);

\draw [-Stealth]  (fp.east) -- (s_linear.west);
\draw [-Stealth]  (s_linear.east) -- (s_ln.west);
\draw [-Stealth]  (s_ln.east) -- (s_hadamard.west);
\draw [-Stealth]  (f.south) -- (s_hadamard.north);
\draw [-Stealth]  (s_hadamard.east) -- (s_conv.west);
\draw [-Stealth]  (coord.east) -- (s_conv.west);
\draw [-Stealth]  (s_conv.east) -- (s_in.west);
\draw [-Stealth]  (s_in.north) -- (s_hadamard2.south);
\draw [-Stealth]  (f.east) -- (s_hadamard2.west);

\draw [-Stealth]  (c_hadamard.east) -- (fusion_conv.west);
\draw [-Stealth]  (s_hadamard2.north) -- (fusion_conv.south);

\draw [-Stealth]  (fusion_conv.east) -- (fusion_in.west);
\draw [-Stealth]  (fusion_in.east) -- (last_conv.west);

\end{tikzpicture}

%% file: tables/coco_testdev.tex
\begin{table}[!t]
\begin{center}
\begin{tabularx}{\linewidth}{|llXXXXXXX|}
\hline
\multicolumn{1}{|l|}{Method}         & \multicolumn{1}{l|}{Backbone}   & \multicolumn{1}{c|}{Input size} & AP & AP$^{50}$ & AP$^{75}$ & AP$^{M}$ & \multicolumn{1}{c|}{AP$^{L}$} & AR \\ \hline
\multicolumn{9}{|c|}{Top-down methods}                                                                                                                            \\ \hline
\multicolumn{1}{|l|}{SBL~\cite{xiao2018simple}} & \multicolumn{1}{l|}{ResNet-152} & \multicolumn{1}{c|}{384$\times$288}    & 73.7 & 91.9 & 81.1 & 70.3 & \multicolumn{1}{l|}{80.0}    & -  \\
\multicolumn{1}{|l|}{HRNet~\cite{sun2019deep}}          & \multicolumn{1}{l|}{HRNet-W32}  & \multicolumn{1}{c|}{384$\times$288}    & 74.9 & 92.5 & 82.8 & 71.3 & \multicolumn{1}{l|}{80.9}    & -  \\ \hline
\multicolumn{9}{|c|}{Bottom-up methods}                                                                                                                           \\ \hline
\multicolumn{1}{|l|}{HrHRNet~\cite{Cheng_2020_CVPR}}    & \multicolumn{1}{l|}{HrHRNet-W32} & \multicolumn{1}{c|}{512}   & 66.4 & 87.5 & 72.8 & 61.2 & \multicolumn{1}{l|}{74.2}    & -  \\
\multicolumn{1}{|l|}{DEKR~\cite{Geng_2021_CVPR}}           & \multicolumn{1}{l|}{HRNet-W32}  & \multicolumn{1}{c|}{512}           & 67.3 & 87.9 & 74.1 & 61.5 & \multicolumn{1}{l|}{76.1}    & 72.4 \\
\multicolumn{1}{|l|}{SWAHR~\cite{Luo_2021_CVPR}}          & \multicolumn{1}{l|}{HrHRNet-W32} & \multicolumn{1}{c|}{512}   & 67.9 & 88.9 & 74.5 & 62.4 & \multicolumn{1}{l|}{75.5}    & - \\ \hline
\multicolumn{9}{|c|}{Single stage methods}                                                                                                                        \\ \hline
\multicolumn{1}{|l|}{FCPose~\cite{Mao_2021_CVPR}}         & \multicolumn{1}{l|}{ResNet-101+FPN} & \multicolumn{1}{c|}{800}    & 65.6 & 87.9 & 72.6 & 62.1 & \multicolumn{1}{l|}{72.3}    & - \\
\multicolumn{1}{|l|}{PETR~\cite{Shi_2022_CVPR}}           & \multicolumn{1}{l|}{ResNet-101}     & \multicolumn{1}{c|}{800} & 68.5 & 90.3 & 76.5 & 62.5 & \multicolumn{1}{l|}{77.0}    & - \\
\multicolumn{1}{|l|}{ED-Pose~\cite{yang2023explicit}}        & \multicolumn{1}{l|}{ResNet-50}      & \multicolumn{1}{c|}{800} & 69.8 & 90.2 & 77.2 & 64.3 & \multicolumn{1}{l|}{77.4}    & - \\
\multicolumn{1}{|l|}{CID~\cite{Wang_2022_CVPR}}            & \multicolumn{1}{l|}{HRNet-W32}      & \multicolumn{1}{c|}{512}    & 68.9 & 89.9 & 76.0 & 63.2 & \multicolumn{1}{l|}{\textbf{77.7}}    & 74.6 \\
\multicolumn{1}{|l|}{CID~\cite{Wang_2022_CVPR}}            & \multicolumn{1}{l|}{HRNet-W48}      & \multicolumn{1}{c|}{640}    & 70.7 & 90.3 & 77.9 & 66.3 & \multicolumn{1}{l|}{\textbf{77.8}}    & 76.4 \\ \hline
\multicolumn{1}{|l|}{BoIR}        & \multicolumn{1}{l|}{HRNet-W32}      & \multicolumn{1}{c|}{512} & \textbf{69.5} & \textbf{90.4} & \textbf{76.9} & \textbf{64.2} & \multicolumn{1}{l|}{77.3} & \textbf{75.3} \\
\multicolumn{1}{|l|}{BoIR}        & \multicolumn{1}{l|}{HRNet-W48}      & \multicolumn{1}{c|}{640} & \textbf{71.2} & \textbf{90.8} & \textbf{78.6} & \textbf{67.0} & \multicolumn{1}{l|}{77.6} & \textbf{77.1} \\ \hline
\end{tabularx}
\end{center}
\caption{Comparison with state-of-the-art methods on COCO \texttt{test-dev} set. Best scores are marked as bold for small (e.g. HRNet-W32) and large (e.g. HRNet-W48) models respectively.}
\label{table:coco_testdev}
\end{table}

%% file: sections/4_experiment.tex
\section{Experiments}
\input{tables/crowdpose_test}
\subsection{Datasets and evaluation metrics}
We evaluated the performance of our approach on four benchmark datasets. 

\vspace{0.1cm}
\noindent \textbf{COCO Keypoint 2017~\cite{10.1007/978-3-319-10602-1_48}} comprises train (57K images), val (5K images), and test-dev (20K images) splits, annotated with 17 keypoints. We use train split for training, and val split for hyperparameter tuning.

\vspace{0.1cm}
\noindent \textbf{CrowdPose~\cite{CrowdPose_CVPR19}} consists of 20K images and 80K instances, annotated with 14 keypoints. Following the evaluation protocol of~\cite{Wang_2022_CVPR}, we use trainval split (12K images, 43.4K instances) for training and test split (8K images, 29K instances) for evaluation.

\vspace{0.1cm}
\noindent \textbf{OCHuman~\cite{zhang2019pose2seg}} is targeted for evaluation on crowded scenes with extreme conditions. 2,500 images are for \texttt{val} set, and 2,231 images are for \texttt{test} set. We evaluate our method following~\cite{jin2020differentiable,Wang_2022_CVPR}.

\vspace{0.1cm}
\noindent \textbf{Evaluation metrics.} We follow COCO evaluation protocol, where AP(Average Precision) and AR(Average Recall) are computed based on OKS(Object Keypoint Similarity) with varying thresholds, including AP (averaged AP), AP${}^{50}$ (AP at OKS=0.5), and AP${}^{75}$ (AP at OKS=0.75).
In case of CrowdPose, we additionally report metrics based on crowd index, including AP${}^{E}$ (easy), AP${}^{M}$ (medium), and AP${}^{H}$ (hard).

\subsection{Implementation details}
\input{tables/coco_val_ochuman}
Our implementation is based on~\cite{Wang_2022_CVPR}.
We use HRNet-W32 and HRNet-W48 as backbone networks and perform hyperparameter tuning with COCO \texttt{val} set results.
We apply AdamW optimizer with initial learning rate 1.0e-3, weight decay 2.5e-2, and cosine learning rate scheduler with 10 warmup epochs, following~\cite{liu2022convnet}.
For COCO, we train the model for 140 epochs on 4 GPUs(RTX 3090 for HRNet-W32, A6000 for HRNet-W48) with AMP, with 20 batch size for each device.
For CrowdPose, similar to~\cite{Wang_2022_CVPR}, we train the model for 310 epochs when training from scratch, while 100 epochs with 1 warmup epoch are applied for transfer learning. Following~\cite{Cheng_2020_CVPR,Geng_2021_CVPR,Wang_2022_CVPR}, we apply single scale test with flipping.

\subsection{Comparison with state-of-the-arts}
\noindent \textbf{Results on COCO datasaet.}
We report COCO \texttt{val} results in Table~\ref{table:coco_val_ochuman}, and \texttt{test-dev} results in Table~\ref{table:coco_testdev}.
Our method outperforms existing state-of-the-art under the same or similar backbone. Our method with HRNet-W32 backbone outperforms CID by 0.8 AP on \texttt{val} and 0.6 AP on \texttt{test-dev}. Similarly, we achieve 0.5 AP improvement on \texttt{test-dev} with HRNet-W48 backbone.
Furthermore, we conducted $t$-test on five COCO \texttt{val} set results from respective methods, and our method achieved statistically significant and consistent improvements over CID with $p$-value $6.4\times 10^{-5}$.

\vspace{1mm}
\noindent \textbf{Results on CrowdPose dataset.}
We compare other methods on CrowdPose \texttt{test} in Table~\ref{table:crowdpose_test}. BoIR is the second best among state-of-the-art methods. Nonetheless, our method suffers from performance drop of 0.7 AP on the HRNet-W32 backbone and 1.1 AP on HRNet-W48 backbone. We speculate that as the model size increases, the model suffers from insufficient amount of training data on CrowdPose, as the performance difference between CID and ED-Pose on CrowdPose is also reversed on COCO. To validate the hypothesis, we introduce finetuning on CrowdPose using the model weights trained on COCO \texttt{train} set. Finetuning strategy is proven to be far more effective, surpassing existing state-of-the-art by 4.5 AP with the HRNet-W32 backbone, and 4.9 AP with HRNet-W48 backbone. For a fair comparison, we additionally conducted the same finetuning strategy on CID, and our method also outperforms the baseline by 0.9 AP.

\begin{table}[]
\begin{center}
\begin{tabular}{|c|c|c|} \hline
Bbox Mask Loss & Bbox Head & AP \\ \hline
           &            & 69.6 \\
\checkmark &            & 70.2 \\
           & \checkmark & 70.4 \\
\checkmark & \checkmark & 70.6 \\ \hline
\end{tabular} \quad
\begin{tabular}{|c|c|c|} \hline
Emb. Loss & Dist. Metric & AP \\ \hline
Contrastive & cosine & 70.3 \\
Contrastive & L2 & 70.2 \\
AE & L2 & 70.6 \\ \hline
\end{tabular}
\end{center}
\caption{Left: Ablation study of Bbox Mask Loss and bbox regression head on COCO \texttt{val} set. Right: Ablation study of embedding loss function and distance metric on COCO \texttt{val} set, where Bbox Mask Loss and bbox head are used.}
\label{table:ablation_main}
\end{table}

\vspace{1mm}
\noindent \textbf{OCHuman results.}
Comparison on OCHuman is summarized in Table~\ref{table:coco_val_ochuman}. Following the protocol in~\cite{jin2020differentiable}, we evaluate the model trained on COCO without finetuning on OCHuman. BoIR outperforms comparative methods on both \texttt{val} and \texttt{test} set by large margin. Therefore, our instance representation learning is effective especially for crowded scenes.

\begin{table}[]
\begin{center}
\begin{tabular}{|l|l|c|c|c|} \hline
Method & Backbone & \# params. (M) & Time (ms) & AP \\ \hline
CID & HRNet-W32 & 29.3 & 86.7 & 69.8 \\
CID & HRNet-W48 & 65.4 & - & - \\
ED-Pose & ResNet-50 & 47.9 & 113.9 & 71.6 \\
ED-Pose & Swin-L & 218.8 & 272.1 & 74.3 \\\hline
BoIR & HRNet-W32 & 31.8 & 110.6 & 70.6 \\
BoIR & HRNet-W48 & 68.9 & 167.3 & 72.5 \\ \hline
\end{tabular}
\end{center}
\caption{Computational cost comparison on COCO \texttt{val} set. Inference time is measured with single RTX 3090 and 1 batch size.}
\label{table:ablation_computation}
\end{table}

\subsection{Ablation study}
\label{subsection:ablation}
We have performed ablative experiments, as illustrated in Table~\ref{table:ablation_main}. 
The effectiveness of Bbox Mask Loss and Bbox Head has been validated by assessing four possible combinations, and the result shows that our proposed methods are useful. We additionally conduct ablative experiments on embedding losses and distance metrics. AE loss turns out to be superior to Contrastive loss. We hypothesize that L2 distance with Gaussian kernel used in AE loss is better suited for keypoint evaluation criteria, as claimed in~\cite{NIPS2017_8edd7215}. We also extensively compare computational cost in Table~\ref{table:ablation_computation}. Our method manages to keep the computational cost within a reasonable extent, compared to ED-Pose. 

For qualitative and visual analysis, we compare our method with CID in Fig.~\ref{fig:qualitative}. The color coding in the figure represents the t-SNE results of the learned features (3 dim). For skateboarding (left), CID missed the border, explaining the inconsistent and less disentangled features. This was evident from the similarity of the color of the border to the background, which also appeared overall noisy. In contrast, BoIR demonstrated clear, consistent, and distinct t-SNE colors for the border, effectively separating it from the background. For right, BoIR further exhibited distinct colors for different individuals, successfully distinguishing between two closely interacting people, where CID failed.

We also visualize the behaviour of our method for overlapping instances in Fig.~\ref{fig:bml_illustration} (b-f). Two boxes A and B in (b), (c), and (d) respectively show the feature similarities of CID from the respective box centers. Similarly, (e) and (f) show the corresponding similarities of BoIR. Our method demonstrates a notably effective separation of features for closely interacting individuals.

%% file: tables/crowdpose_test.tex
\begin{table*}[!t]
\begin{center}
\begin{tabularx}{\linewidth}{|llXXXXXXX|}
\hline
\multicolumn{1}{|l|}{Method}         & \multicolumn{1}{l|}{Backbone}   & \multicolumn{1}{c|}{Input size} & AP & AP$^{50}$ & \multicolumn{1}{c|}{AP$^{75}$} & AP$^{E}$ & AP$^{M}$ & AP$^{H}$ \\ \hline
\multicolumn{9}{|c|}{Top-down methods}                                                                                                                            \\ \hline
\multicolumn{1}{|l|}{SBL~\cite{xiao2018simple}} & \multicolumn{1}{l|}{ResNet-101} & \multicolumn{1}{c|}{-}    & 60.8 & 81.4 & \multicolumn{1}{l|}{65.7} & 71.4 & 61.2 & 51.2  \\
\multicolumn{1}{|l|}{SPPE~\cite{CrowdPose_CVPR19}}          & \multicolumn{1}{l|}{ResNet-101}  & \multicolumn{1}{c|}{320$\times$ 256}    & 66.0 & 84.2 & \multicolumn{1}{l|}{71.5} & 75.5 & 66.3 & 57.4 \\ \hline
\multicolumn{9}{|c|}{Bottom-up methods}                                                                                                                           \\ \hline
\multicolumn{1}{|l|}{HrHRNet~\cite{Cheng_2020_CVPR}}    & \multicolumn{1}{l|}{HrHRNet-W48} & \multicolumn{1}{c|}{640}   & 65.9 & 86.4 & \multicolumn{1}{l|}{70.6} & 73.3 & 66.5 & 57.9  \\
\multicolumn{1}{|l|}{DEKR~\cite{Geng_2021_CVPR}}           & \multicolumn{1}{l|}{HrHRNet-W32}  & \multicolumn{1}{c|}{512}           & 65.7 & 85.7 & \multicolumn{1}{l|}{70.4} & 73.0 & 66.4 & 57.5 \\
\multicolumn{1}{|l|}{SWAHR~\cite{Luo_2021_CVPR}}          & \multicolumn{1}{l|}{HrHRNet-W48} & \multicolumn{1}{c|}{640}   & 71.6 & 88.5 & \multicolumn{1}{l|}{77.6} & 78.9 & 72.4 & 63.0 \\ \hline
\multicolumn{9}{|c|}{Single stage methods}                                                                                                                        \\ \hline
\multicolumn{1}{|l|}{PETR~\cite{Shi_2022_CVPR}}           & \multicolumn{1}{l|}{Swin-L}     & \multicolumn{1}{c|}{800} & 71.6 & 90.4 & \multicolumn{1}{l|}{78.3} & 77.3 & 72.0 & 65.8 \\
\multicolumn{1}{|l|}{ED-Pose~\cite{yang2023explicit}}        & \multicolumn{1}{l|}{ResNet-50}      & \multicolumn{1}{c|}{800} & 69.9 & 88.6 & \multicolumn{1}{l|}{75.8} & 77.7 & 70.6 & 60.9 \\
\multicolumn{1}{|l|}{CID~\cite{Wang_2022_CVPR}}            & \multicolumn{1}{l|}{HRNet-W32}      & \multicolumn{1}{c|}{512}    & 71.3 & 90.6 & \multicolumn{1}{l|}{76.6} & 77.4 & 72.1 & 63.9 \\
\multicolumn{1}{|l|}{CID~\cite{Wang_2022_CVPR}}            & \multicolumn{1}{l|}{HRNet-W48}      & \multicolumn{1}{c|}{640}    & 72.3 & 90.8 & \multicolumn{1}{l|}{77.9} & 78.7 & 73.0 & 64.8 \\
\multicolumn{1}{|l|}{CID${}^{*}$~\cite{Wang_2022_CVPR}}    & \multicolumn{1}{l|}{HRNet-W32}      & \multicolumn{1}{c|}{512}    & 74.9 & 91.8 & \multicolumn{1}{l|}{81.0} & 82.0 & 75.8 & 66.3 \\ \hline
\multicolumn{1}{|l|}{BoIR}        & \multicolumn{1}{l|}{HRNet-W32} & \multicolumn{1}{c|}{512} & 70.6 & 89.9 & \multicolumn{1}{l|}{76.5} & 77.1 & 71.2 & 63.0 \\ 
\multicolumn{1}{|l|}{BoIR}        & \multicolumn{1}{l|}{HRNet-W48} & \multicolumn{1}{c|}{640} & 71.2 & 90.3 & \multicolumn{1}{l|}{76.7} & 77.8 & 71.8 & 63.5 \\
\multicolumn{1}{|l|}{BoIR${}^{*}$}        & \multicolumn{1}{l|}{HRNet-W32} & \multicolumn{1}{c|}{512} & \textbf{75.8} & \textbf{92.2} & \multicolumn{1}{l|}{\textbf{82.3}} & \textbf{82.3} & \textbf{76.5} & \textbf{67.5} \\ 
\multicolumn{1}{|l|}{BoIR${}^{*}$}        & \multicolumn{1}{l|}{HRNet-W48} & \multicolumn{1}{c|}{640} & \textbf{77.2} & \textbf{92.4} & \multicolumn{1}{l|}{\textbf{83.5}} & \textbf{82.7} & \textbf{78.1} & \textbf{69.8} \\ \hline
\end{tabularx}
\end{center}
\caption{Comparison with state-of-the-art methods on CrowdPose \texttt{test} set. Best scores are marked as bold for small(e.g. HRNet-W32) and large(e.g. HRNet-W48) models respectively. Models with $*$ are trained on COCO and finetuned on CrowdPose.}
\label{table:crowdpose_test}
\end{table*}

%% file: tables/coco_val_ochuman.tex
\begin{table}[!t]
\begin{center}
\begin{tabular}{|l|l|cc|cc|cc|}
\hline
\multirow{2}{*}{Method}         & \multirow{2}{*}{Backbone} & \multicolumn{2}{c|}{COCO \texttt{val}} & \multicolumn{2}{c|}{OCHuman \texttt{val}} & \multicolumn{2}{c|}{OCHuman \texttt{test}} \\ \cline{3-8}
& & AP & AR & AP & AR & AP & AR \\\hline
\multicolumn{1}{|l|}{DEKR~\cite{Geng_2021_CVPR}}  & \multicolumn{1}{l|}{HRNet-W32}  & 68.0 & 73.0 & 37.9 & - & 36.5 & - \\
\multicolumn{1}{|l|}{DEKR~\cite{Geng_2021_CVPR}}  & \multicolumn{1}{l|}{HRNet-W48}  & 71.0 & 76.0 & - & - & - & - \\ \hline
\multicolumn{1}{|l|}{CID~\cite{Wang_2022_CVPR}}  & \multicolumn{1}{l|}{HRNet-W32}  & 69.8 & 75.4 & 44.9 & - & 44.0 & - \\
\multicolumn{1}{|l|}{CID~\cite{Wang_2022_CVPR}}  & \multicolumn{1}{l|}{HRNet-W48}  & - & - & 46.1 & - & 45.0 & - \\ \hline
\multicolumn{1}{|l|}{BoIR}        & \multicolumn{1}{l|}{HRNet-W32}      & \textbf{70.6} & \textbf{76.3} & \textbf{47.4} & \textbf{80.1} & \textbf{47.0} & \textbf{80.3} \\
\multicolumn{1}{|l|}{BoIR}        & \multicolumn{1}{l|}{HRNet-W48}      & \textbf{72.5} & \textbf{78.3} & \textbf{49.4} & \textbf{80.8} & \textbf{48.5} & \textbf{80.7} \\ \hline
\end{tabular}
\end{center}
\caption{Comparison with state-of-the-art methods on COCO \texttt{val} and OCHuman \texttt{val}, \texttt{test} set. OCHuman performance is evaluated with COCO pretrained model without finetuning.}
\label{table:coco_val_ochuman}
\end{table}

%% file: sections/5_conclusion.tex
\section{Conclusion}
\input{figures/3_emb_vis.tex}
This paper proposes a new multi-person pose estimation method using bounding box-supervised instance representation learning, called BoIR. It provides rich spatial supervision, utilizing embedding similarity as a soft mask for positive sampling, and the background region as a negative sample. It also incorporates auxiliary tasks for richer multi-task learning, without additional computation cost during inference. Our instance embedding can effectively disentangle instances in crowded scenes, surpassing comparable state-of-the-art methods on multiple human pose estimation benchmarks. Despite notable performance improvement with transfer learning, effective representation learning on small training data is a remaining issue, and we plan to mitigate the limitation as future work. 
Potential future work also involves enhancing auxiliary supervision by incorporating additional tasks, \eg action recognition, and leveraging image captions within the framework of multi-modal contrastive training.


%% file: figures/3_emb_vis.tex
\begin{figure}
\begin{center}
\begin{tabular}{c@{\hspace{0.3em}}c@{\hspace{0.3em}}c|c@{\hspace{0.3em}}c}
 & Keypoint & t-SNE & Keypoint & t-SNE \\
\rotatebox[origin=c]{90}{CID} & \raisebox{-.5\height}{\includegraphics[width=0.22\linewidth]{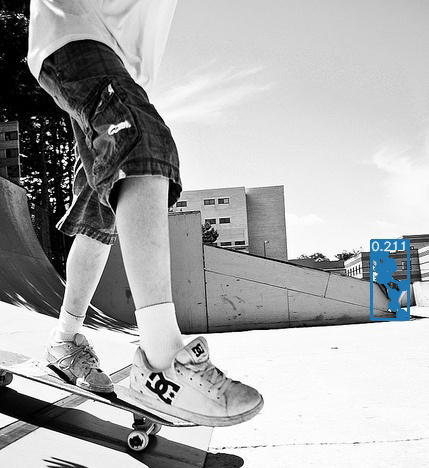}} & \raisebox{-.5\height}{\includegraphics[width=0.22\linewidth]{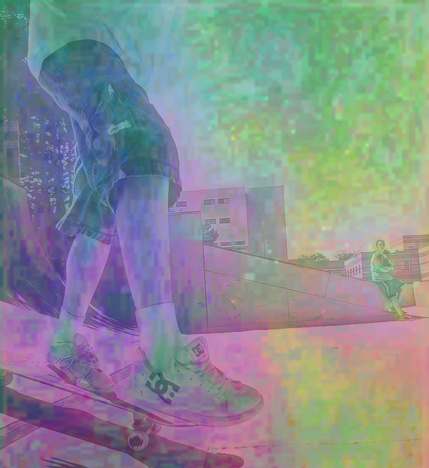}} & \raisebox{-.5\height}{\includegraphics[width=0.2\linewidth]{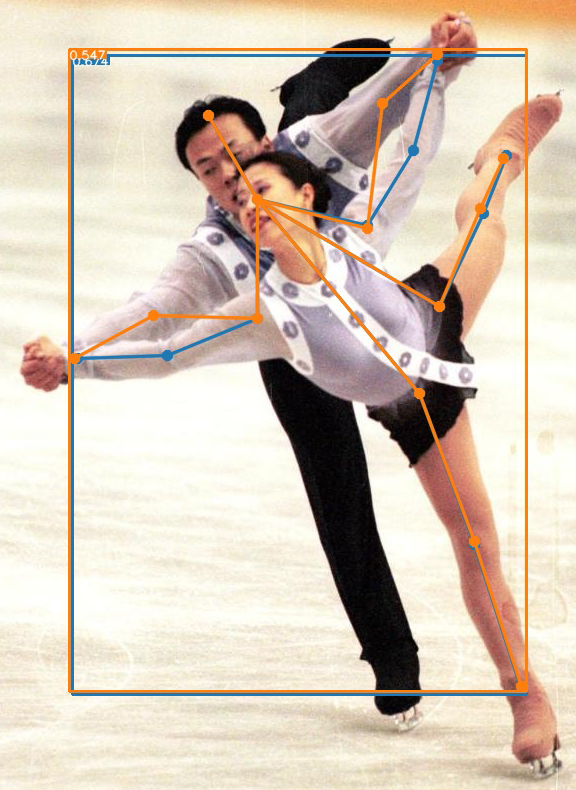}} & \raisebox{-.5\height}{\includegraphics[width=0.2\linewidth]{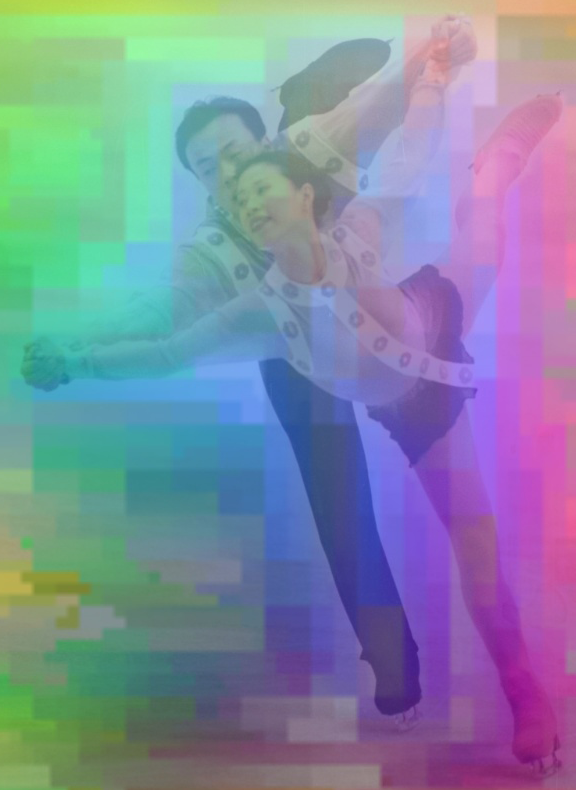}} \\
\rotatebox[origin=c]{90}{BoIR} & \raisebox{-.5\height}{\includegraphics[width=0.22\linewidth]{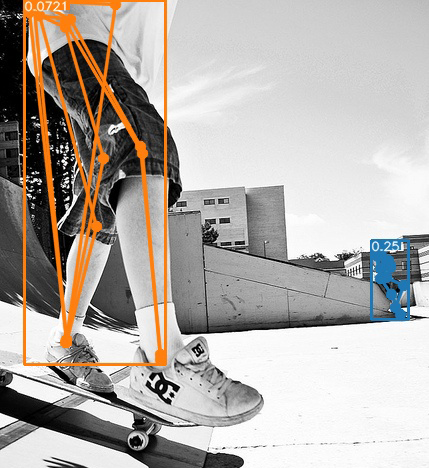}} & \raisebox{-.5\height}{\includegraphics[width=0.22\linewidth]{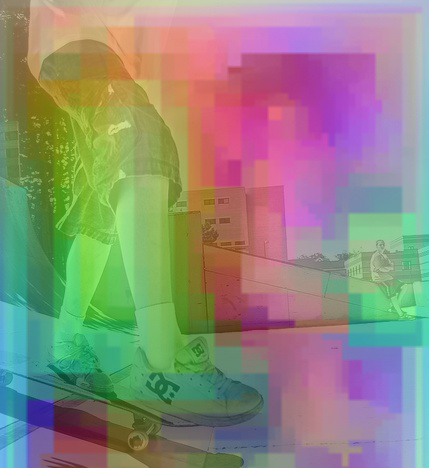}} & \raisebox{-.5\height}{\includegraphics[width=0.2\linewidth]{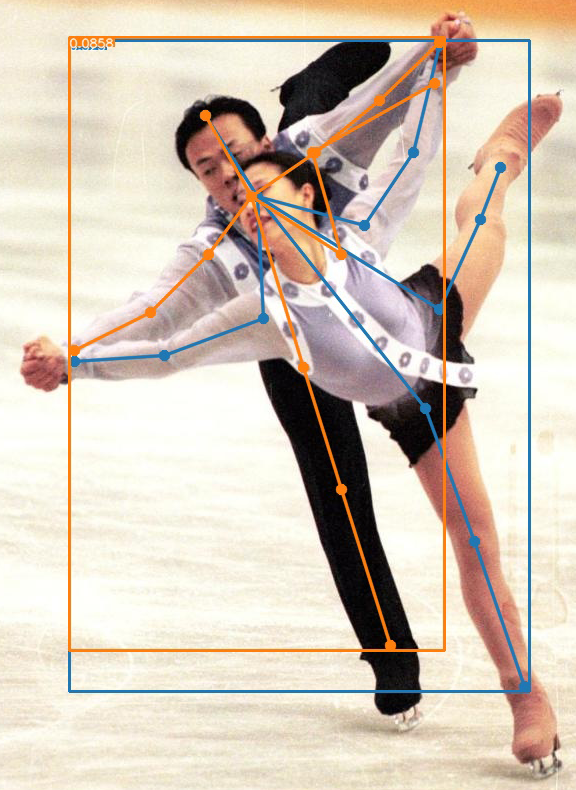}} & \raisebox{-.5\height}{\includegraphics[width=0.2\linewidth]{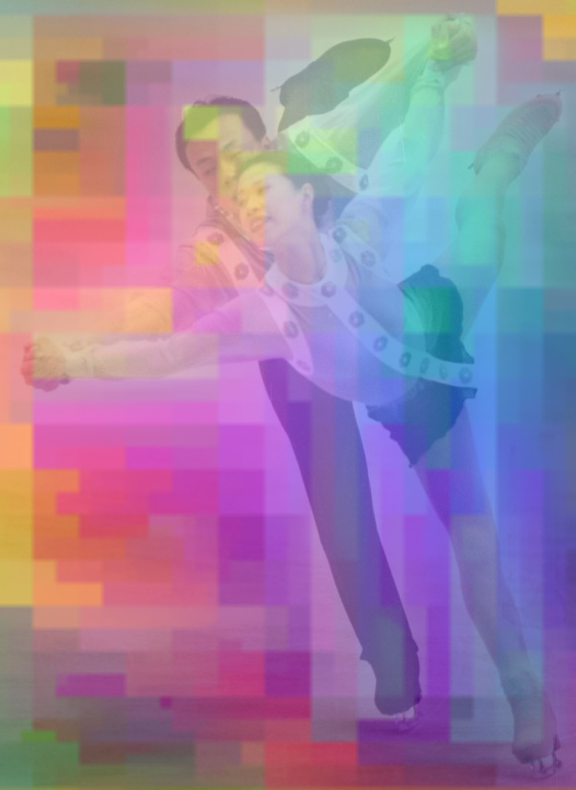}} \\ 
\end{tabular}
\end{center}
\caption{Example outcomes using our approach. The image on the left is from the COCO \texttt{val} set, while the image on the right is from the CrowdPose \texttt{test} set. We employed t-SNE, running it for 250 iterations, on the output backbone feature, with three output dimensions per pixel, corresponding directly to normalized RGB values.
}
\label{fig:qualitative}
\end{figure}

%% file: sections/supplementary.tex
In this supplementary material, we provide further details about 1) Bbox Mask Loss 2) architecture composition 3) additional experiments 4) visualization and comparative analysis with CID.

\section{Bbox Mask Loss}
From the instance embedding map $e$, we first apply L2 normalization on $e$.
Then, we sample instance embedding $p$ from $e$ and compute respective loss terms as following:
\begin{equation}
\mathcal{L}^{in}_{pull} = \frac{1}{DN}\sum_{i=1}^{N}\sum_{d=1}^{D}\left(p_{(i,d)} - \bar{p}_{(i,d)}\right)^2
\label{eq:sup1}
\end{equation}
\begin{equation}
\mathcal{L}^{out}_{push} = \frac{1}{N}\sum_{i=1}^{N} \exp \left\{-\frac{\beta}{D} \sum_{d=1}^{D} \left(p_{(i,d)} - \bar{p}^{c}_{(i,d)}\right)^2 \right\}
\label{eq:sup2}
\end{equation}
\begin{equation}
\mathcal{L}^{inst}_{push} = \frac{1}{\frac{N(N-1)}{2}}\sum_{i=1}^{N}\sum_{j>i}^{N} \exp \left\{-\frac{\beta}{D} \sum_{d=1}^{D} \left(p_{(i,d)} - p_{(j,d)}\right)^2 \right\}
\label{eq:sup3}
\end{equation}

$D$ is the dimension of the instance embedding, $N$ is the number of ground-truth instances in an image, and $i,j$ represent the instance indices. $\beta=\frac{1}{2\sigma^2}$ is a scaling coefficient for the Gaussian kernel proposed in Associative Embedding~\cite{NIPS2017_8edd7215}.

\section{Architecture Composition}
In the case of 512x512 input size, output heatmap size is set to 128x128. In the case of 640x640 input size, output heatmap size is 160x160. HRNet-W32 backbone outputs 480 channels, due to concatenation of all block outputs. Similarly, HRNet-W48 backbone outputs 720 channels. 

In the case of auxiliary task heads, 1 residual block and 1 convolution layer are applied.
Residual block receives 256 input channels and outputs 128 channels. The final convolution layer outputs task-specific output channels. In case of bottom-up keypoint head, it is the number of keypoints(17 in COCO, 14 in CrowdPose). In case of the bounding box head, it outputs 4 channels(left, top, right, bottom distance). In case of embedding head, it is $D$. All convolution layers in the auxiliary task head have a 3x3 kernel size.

In case of instance-wise keypoint regression head, 64 hidden channel size is applied for HRNet-W32 backbone. 96 hidden channel size is used for HRNet-W48 backbone.

\section{Additional Experiments}
\input{tables/coco_val_full}
We report full comparative evaluation results on COCO \texttt{val} set on Table~\ref{table:coco_val_full}. CID paper does not report full results, so we report the scores using the provided trained model weights. Since HRNet-W48 backbone model is not available, CID's HRNet-W48 results are not reported. BoIR outperforms all comparative stae-of-the-arts except for HRNet-W48 backbone on $AP^{L}$. Our method with HRNet-W32 backbone outperforms CID by 0.8 AP. Our method with HRNet-W48 even outperforms ED-Pose by 0.9 AP.

\input{tables/beta_dim_ablation}
We report ablation experiments on AE's Gaussian kernel scaling coefficient $\beta$ and embedding dimension $D$ on COCO \texttt{val} set on Table~\ref{table:ablation_beta_dim}. For fast training and simplicity, we use Bbox Mask Loss and do not use bbox head during experiment. In case of $\beta$, 10 was the best among the options. In case of $D$, changing the embedding dimension shows little performance difference. We conjecture that AE loss for higher dimensions needs more refinement to benefit high dimensional representation. The primary cause is L2 normalization over embedding dimension before loss computation, which significantly drops the loss scale and floating point precision compared to the original AE loss formulation. Simply removing the normalization would cause unstable training with AMP, so further research is required to improve the current framework.

\section{Visualization}
\begin{figure}
\begin{center}
\begin{tabular}{ccc|cc}
 & CID & BoIR & CID & BoIR \\
GT    & \raisebox{-.5\height}{\includegraphics[width=0.2\linewidth]{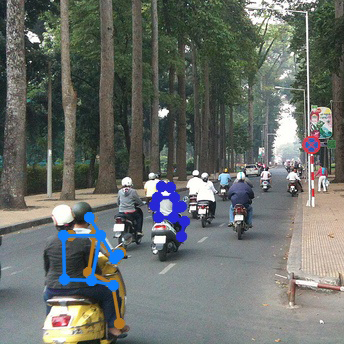}} & \raisebox{-.5\height}{\includegraphics[width=0.2\linewidth]{images/supp/gt_000000314251.png}} & \raisebox{-.5\height}{\includegraphics[width=0.2\linewidth]{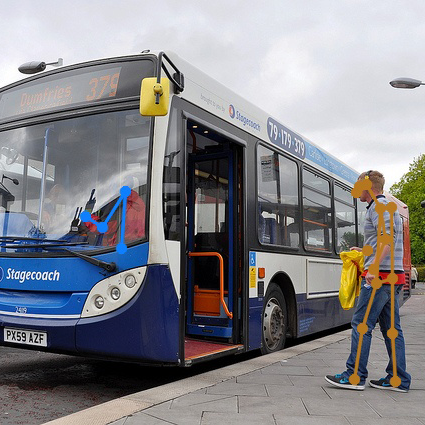}} & \raisebox{-.5\height}{\includegraphics[width=0.2\linewidth]{images/supp/gt_000000333402.png}} \\ 
Pred  & \raisebox{-.5\height}{\includegraphics[width=0.2\linewidth]{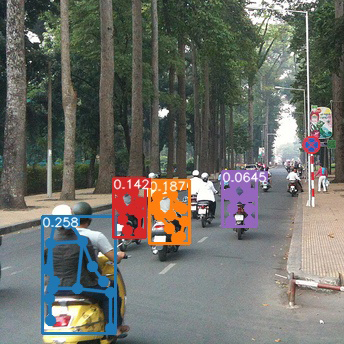}} & \raisebox{-.5\height}{\includegraphics[width=0.2\linewidth]{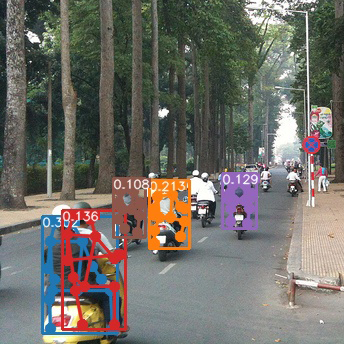}} & \raisebox{-.5\height}{\includegraphics[width=0.2\linewidth]{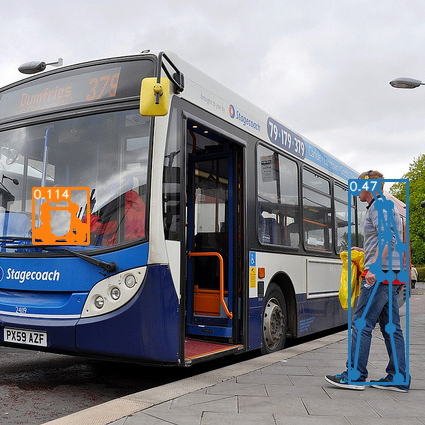}} & \raisebox{-.5\height}{\includegraphics[width=0.2\linewidth]{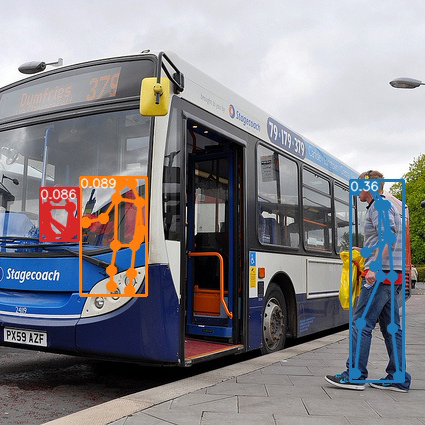}} \\ 
Center& \raisebox{-.5\height}{\includegraphics[width=0.2\linewidth]{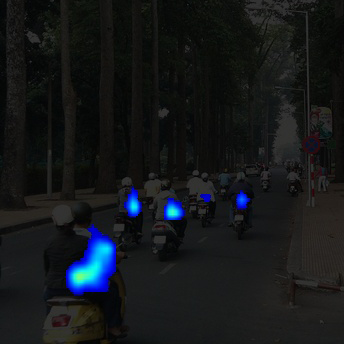}} & \raisebox{-.5\height}{\includegraphics[width=0.2\linewidth]{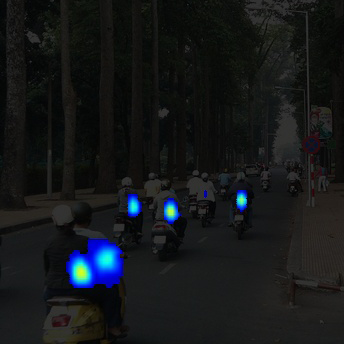}} & \raisebox{-.5\height}{\includegraphics[width=0.2\linewidth]{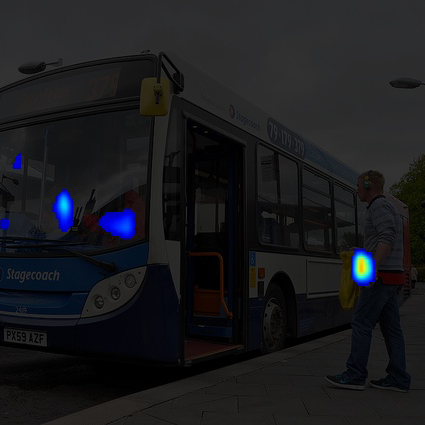}} & \raisebox{-.5\height}{\includegraphics[width=0.2\linewidth]{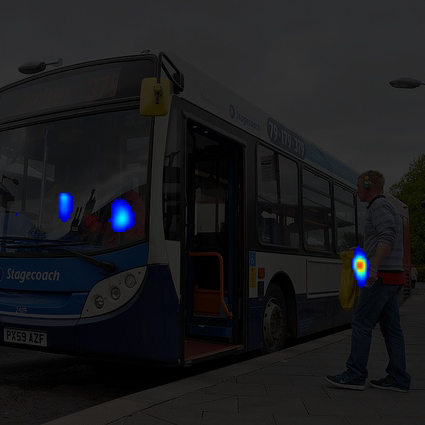}} \\ 
t-SNE & \raisebox{-.5\height}{\includegraphics[width=0.2\linewidth]{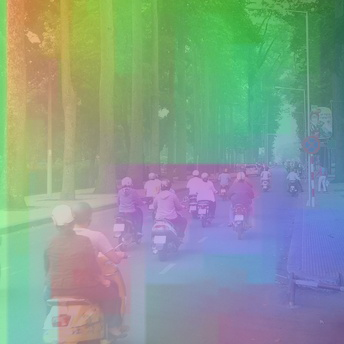}} & \raisebox{-.5\height}{\includegraphics[width=0.2\linewidth]{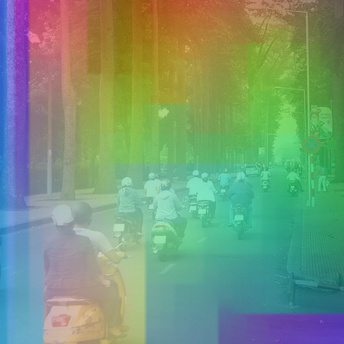}} & \raisebox{-.5\height}{\includegraphics[width=0.2\linewidth]{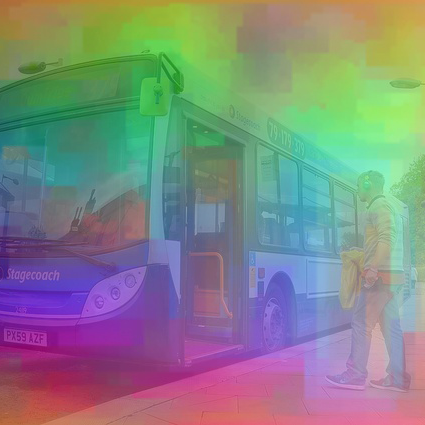}} & \raisebox{-.5\height}{\includegraphics[width=0.2\linewidth]{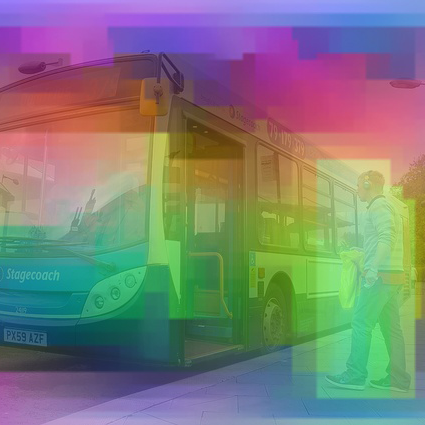}} \\ 
Sim   & \raisebox{-.5\height}{\includegraphics[width=0.2\linewidth]{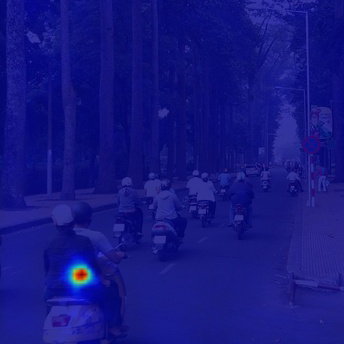}} & \raisebox{-.5\height}{\includegraphics[width=0.2\linewidth]{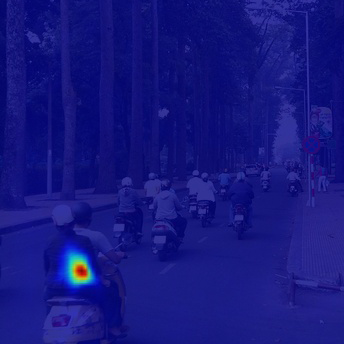}} & \raisebox{-.5\height}{\includegraphics[width=0.2\linewidth]{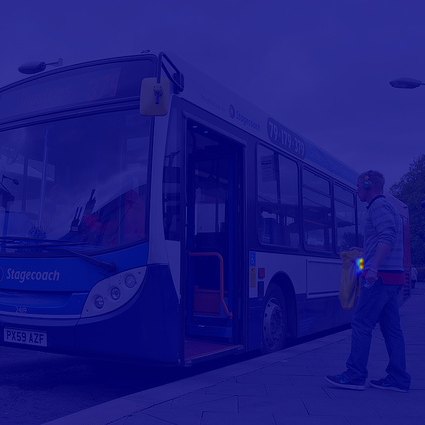}} & \raisebox{-.5\height}{\includegraphics[width=0.2\linewidth]{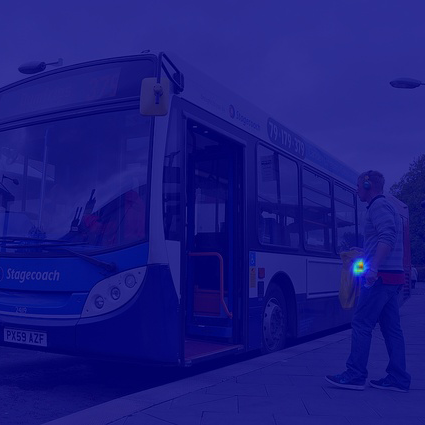}} \\ 
\end{tabular}
\end{center}
\caption{Comparative visualization on COCO \texttt{val} set.}
\label{fig:supp_vis_coco}
\end{figure}

\begin{figure}
\begin{center}
\begin{tabular}{ccc|cc}
 & CID & BoIR & CID & BoIR \\
GT    & \raisebox{-.5\height}{\includegraphics[width=0.2\linewidth]{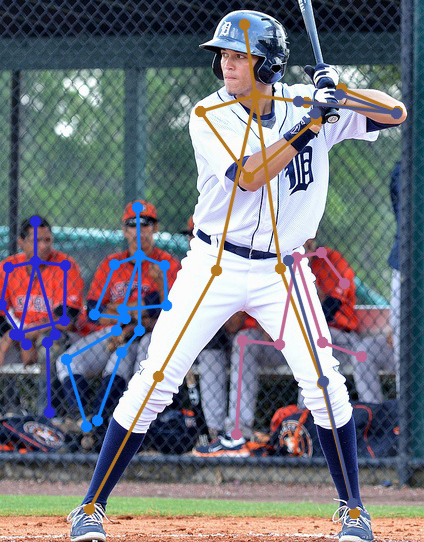}} & \raisebox{-.5\height}{\includegraphics[width=0.2\linewidth]{images/supp/gt_105713.png}} & \raisebox{-.5\height}{\includegraphics[width=0.2\linewidth]{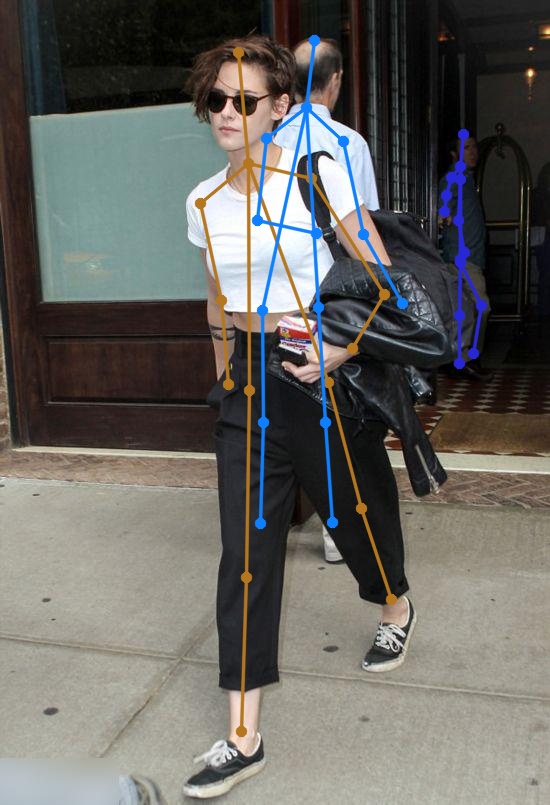}} & \raisebox{-.5\height}{\includegraphics[width=0.2\linewidth]{images/supp/gt_111878.jpg}} \\ 
Pred  & \raisebox{-.5\height}{\includegraphics[width=0.2\linewidth]{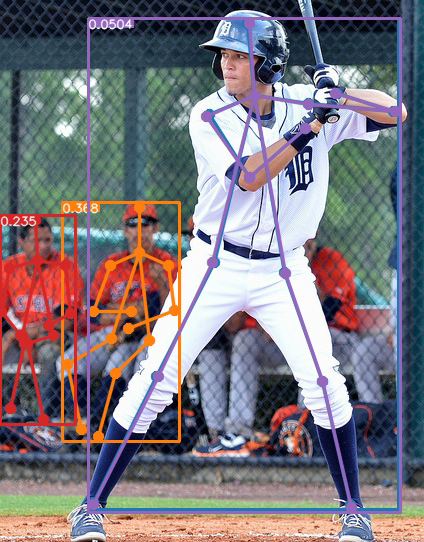}} & \raisebox{-.5\height}{\includegraphics[width=0.2\linewidth]{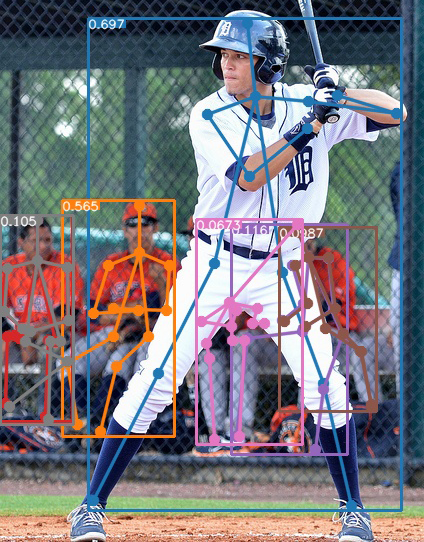}} & \raisebox{-.5\height}{\includegraphics[width=0.2\linewidth]{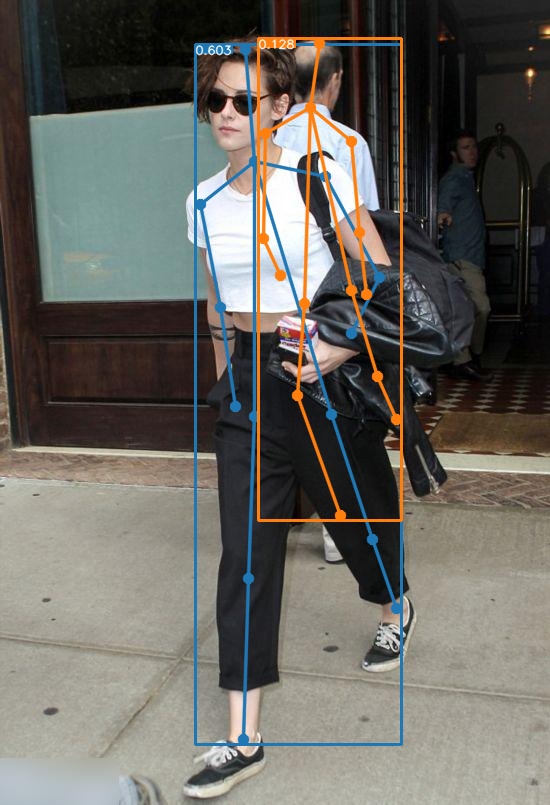}} & \raisebox{-.5\height}{\includegraphics[width=0.2\linewidth]{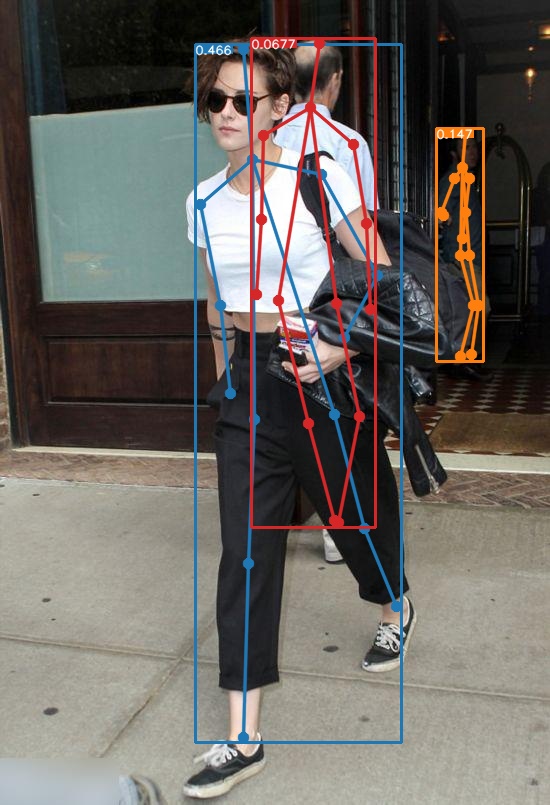}} \\ 
Center& \raisebox{-.5\height}{\includegraphics[width=0.2\linewidth]{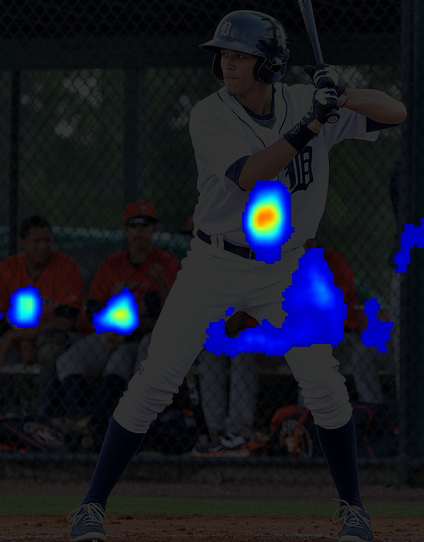}} & \raisebox{-.5\height}{\includegraphics[width=0.2\linewidth]{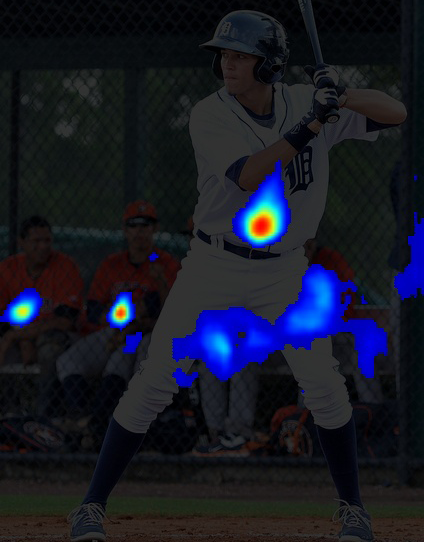}} & \raisebox{-.5\height}{\includegraphics[width=0.2\linewidth]{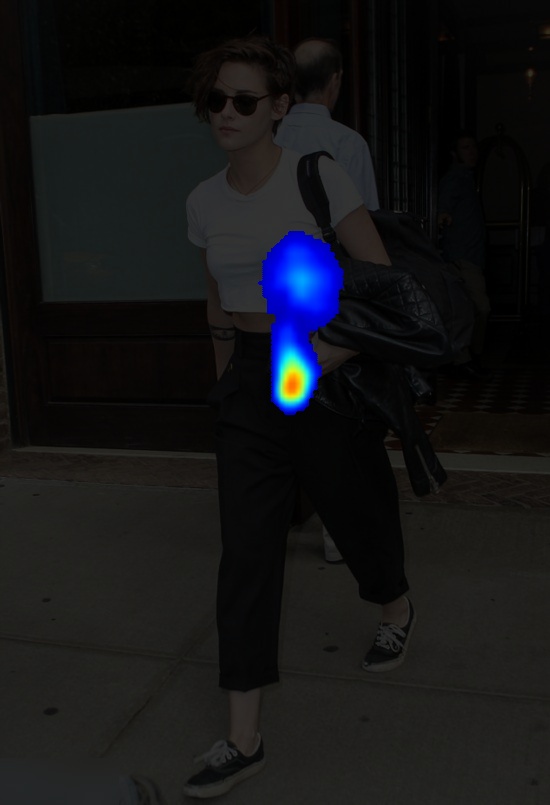}} & \raisebox{-.5\height}{\includegraphics[width=0.2\linewidth]{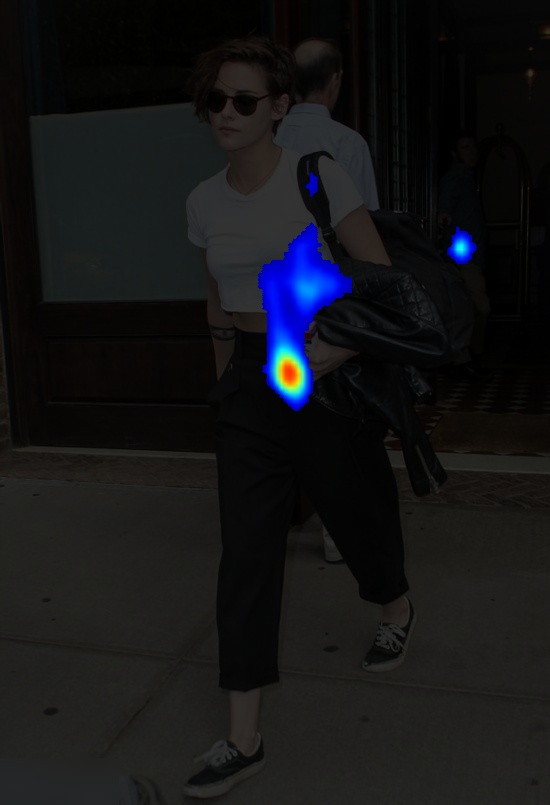}} \\ 
t-SNE & \raisebox{-.5\height}{\includegraphics[width=0.2\linewidth]{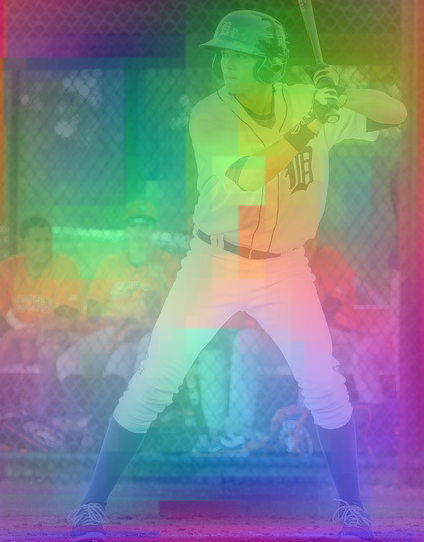}} & \raisebox{-.5\height}{\includegraphics[width=0.2\linewidth]{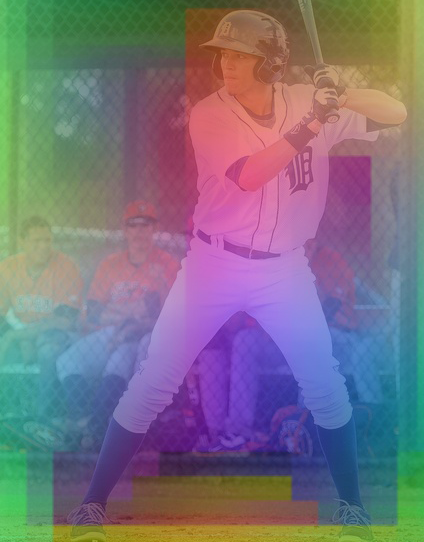}} & \raisebox{-.5\height}{\includegraphics[width=0.2\linewidth]{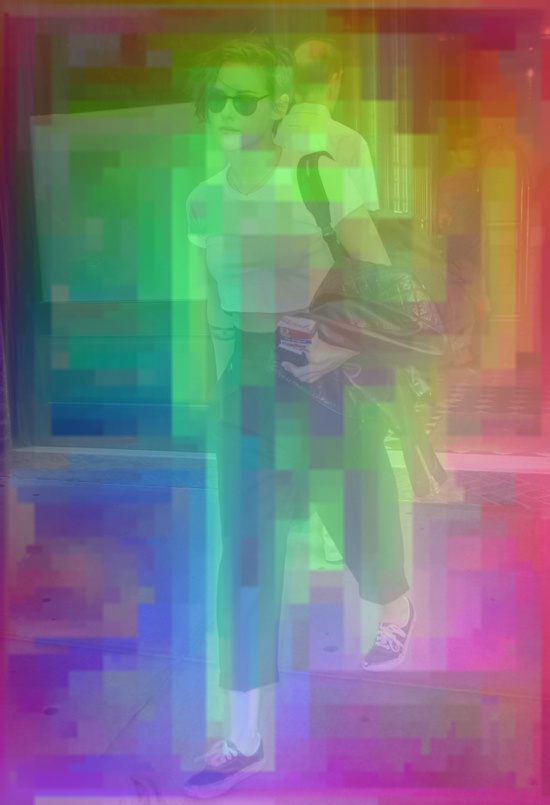}} & \raisebox{-.5\height}{\includegraphics[width=0.2\linewidth]{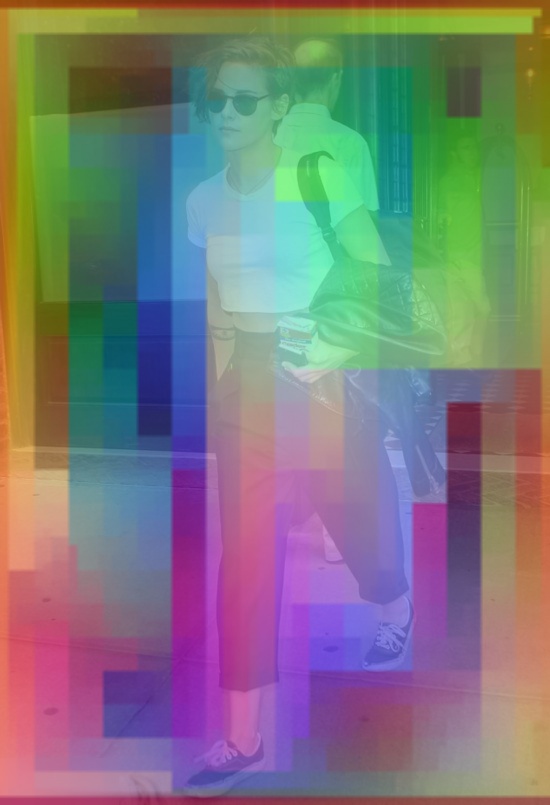}} \\ 
Sim   & \raisebox{-.5\height}{\includegraphics[width=0.2\linewidth]{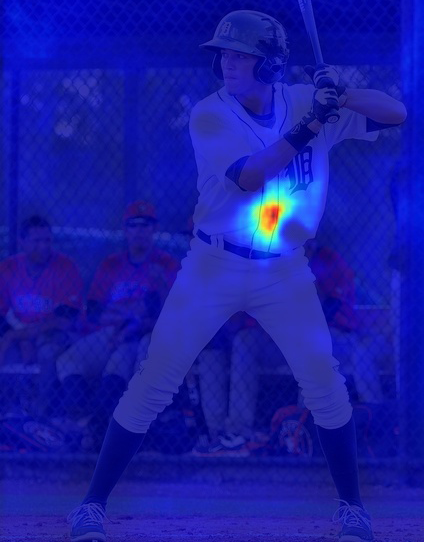}} & \raisebox{-.5\height}{\includegraphics[width=0.2\linewidth]{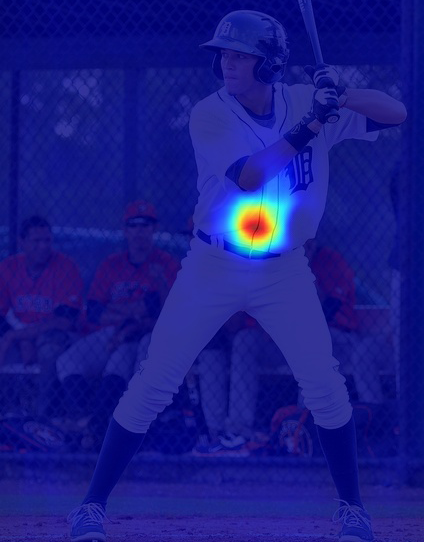}} & \raisebox{-.5\height}{\includegraphics[width=0.2\linewidth]{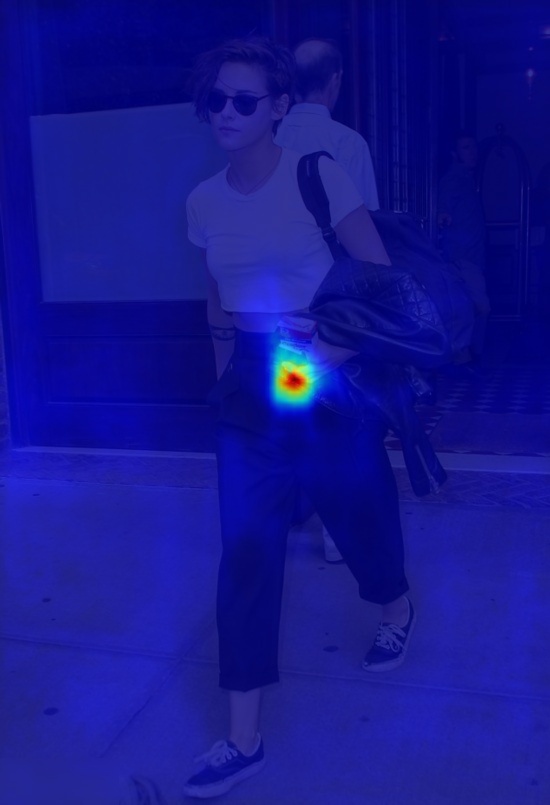}} & \raisebox{-.5\height}{\includegraphics[width=0.2\linewidth]{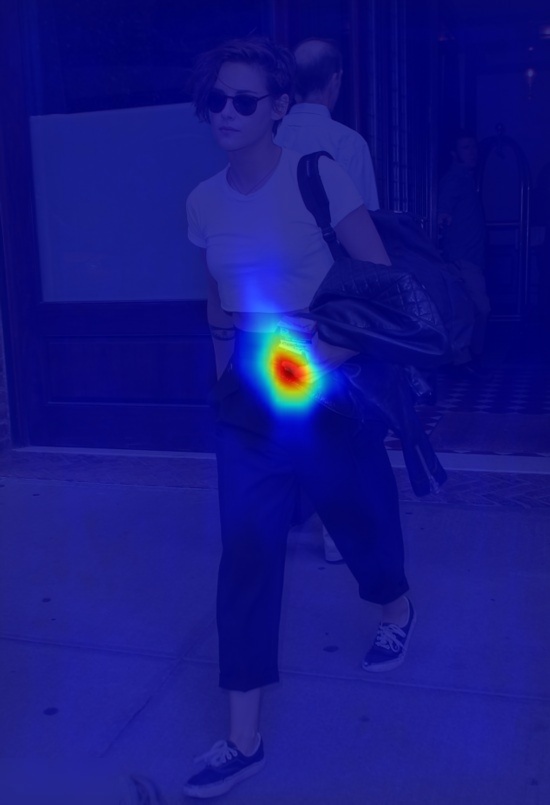}} \\ 
\end{tabular}
\end{center}
\caption{Comparative visualization on CrowdPose \texttt{test} set.}
\label{fig:supp_vis_crowdpose}
\end{figure}

We provide extensive outputs of our model in Fig.~\ref{fig:supp_vis_coco} and Fig.~\ref{fig:supp_vis_crowdpose}. We visualize keypoint prediction outputs along with instance center heatmap, t-SNE of backbone output feature, and feature similarity between top-1 confident instance parameter and the entire feature map. t-SNE is applied on the output backbone feature for 250 iterations with 3 output dimensions per pixel, which directly corresponds to normalized RGB values. Instance similarity is measured by computing the L2 distance between the top-1 confident instance's parameter and the feature map, and then applying a Gaussian Kernel.

\begin{figure}
\begin{center}
\begin{tabular}{ccc|cc}
 & CID & BoIR & CID & BoIR \\
GT    & \raisebox{-.5\height}{\includegraphics[width=0.2\linewidth]{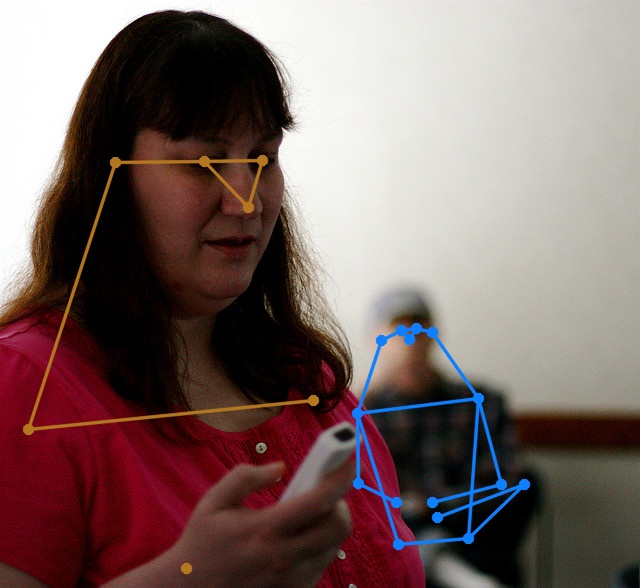}} & \raisebox{-.5\height}{\includegraphics[width=0.2\linewidth]{images/supp/gt_000000383838.jpg}} & \raisebox{-.5\height}{\includegraphics[width=0.2\linewidth]{images/supp/gt_000000235241.jpg}} & \raisebox{-.5\height}{\includegraphics[width=0.2\linewidth]{images/supp/gt_000000235241.jpg}} \\ 
Pred  & \raisebox{-.5\height}{\includegraphics[width=0.2\linewidth]{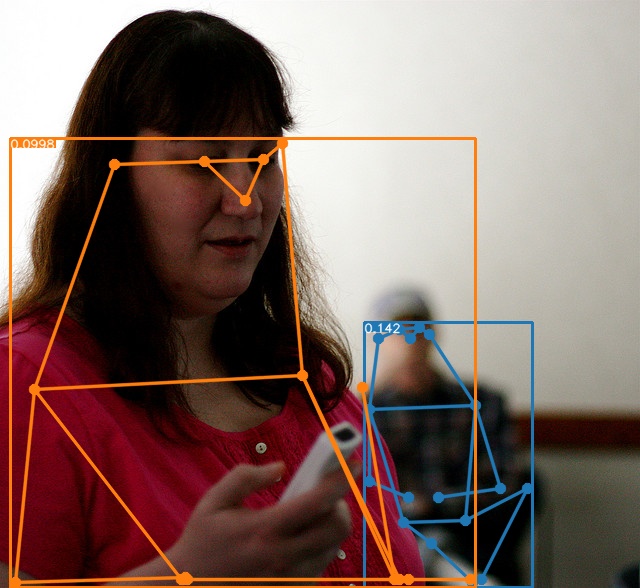}} & \raisebox{-.5\height}{\includegraphics[width=0.2\linewidth]{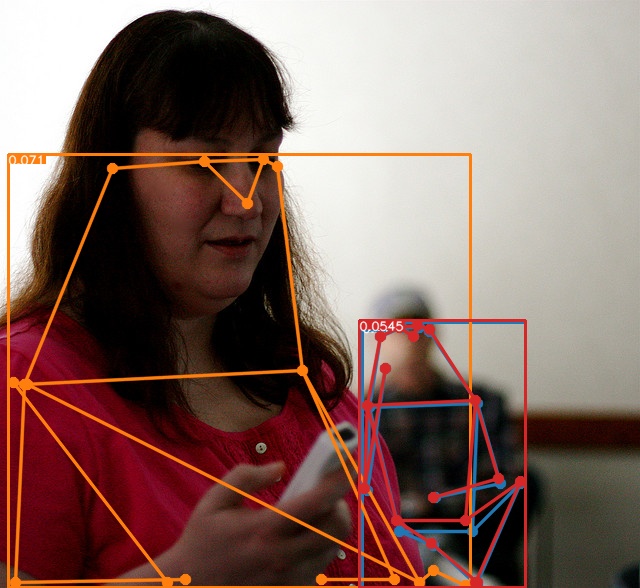}} & \raisebox{-.5\height}{\includegraphics[width=0.2\linewidth]{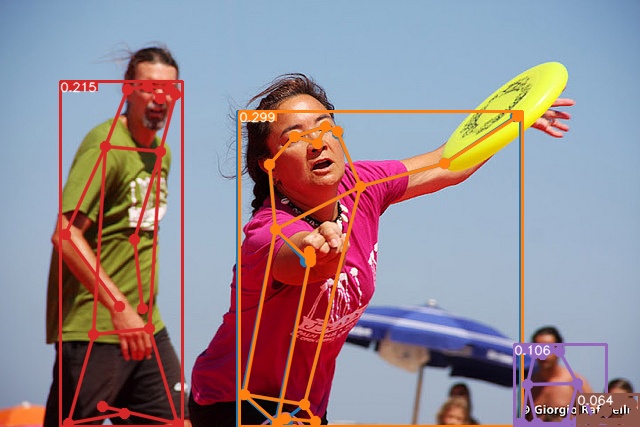}} & \raisebox{-.5\height}{\includegraphics[width=0.2\linewidth]{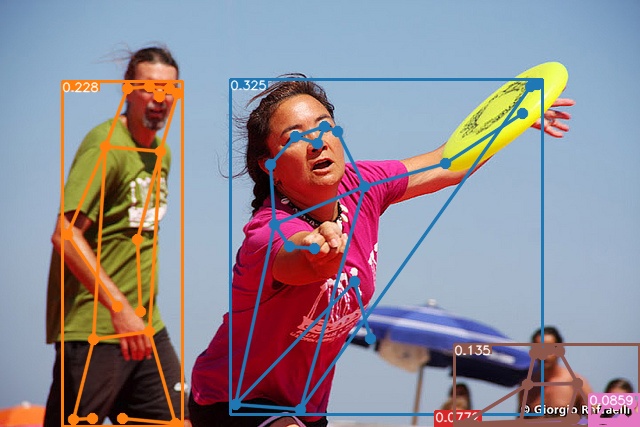}} \\ 
Center& \raisebox{-.5\height}{\includegraphics[width=0.2\linewidth]{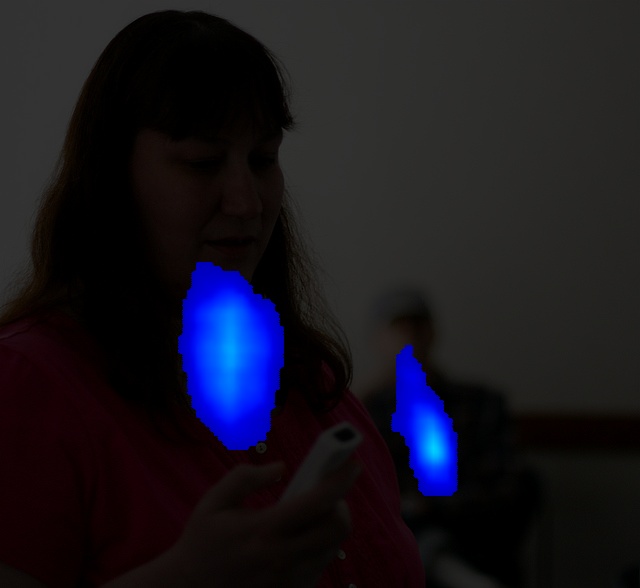}} & \raisebox{-.5\height}{\includegraphics[width=0.2\linewidth]{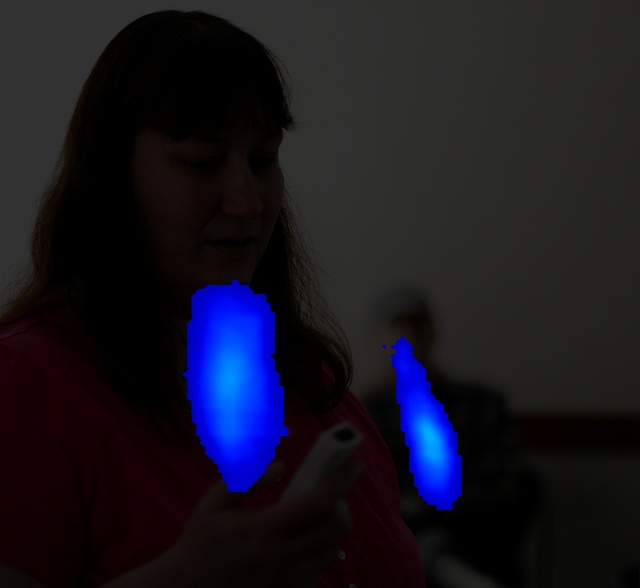}} & \raisebox{-.5\height}{\includegraphics[width=0.2\linewidth]{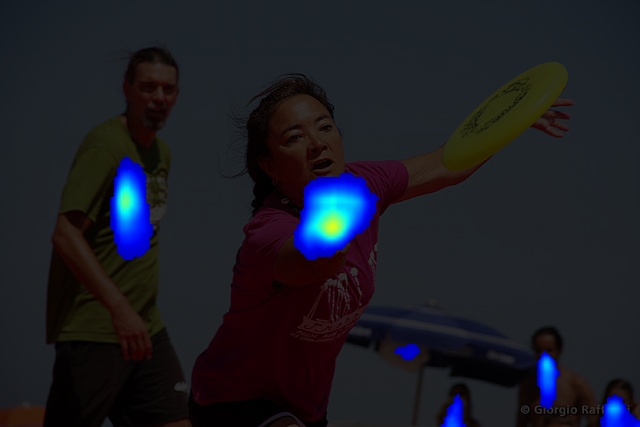}} & \raisebox{-.5\height}{\includegraphics[width=0.2\linewidth]{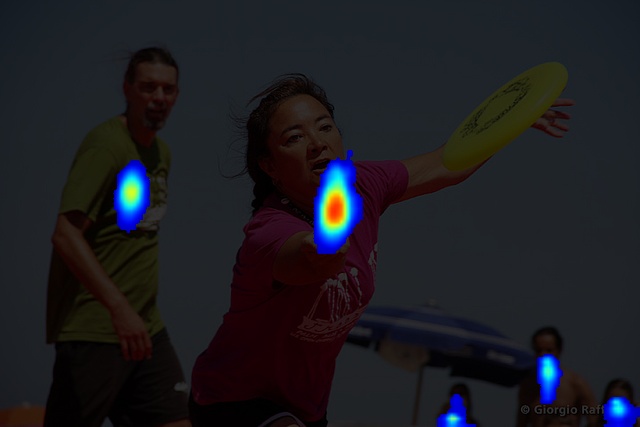}} \\ 
t-SNE & \raisebox{-.5\height}{\includegraphics[width=0.2\linewidth]{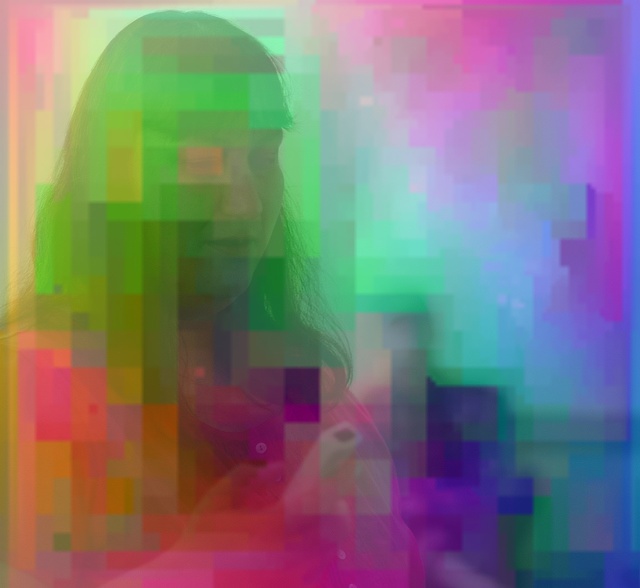}} & \raisebox{-.5\height}{\includegraphics[width=0.2\linewidth]{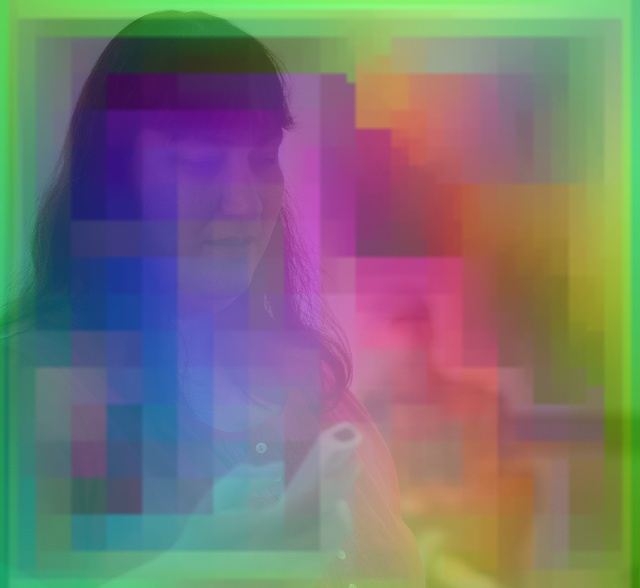}} & \raisebox{-.5\height}{\includegraphics[width=0.2\linewidth]{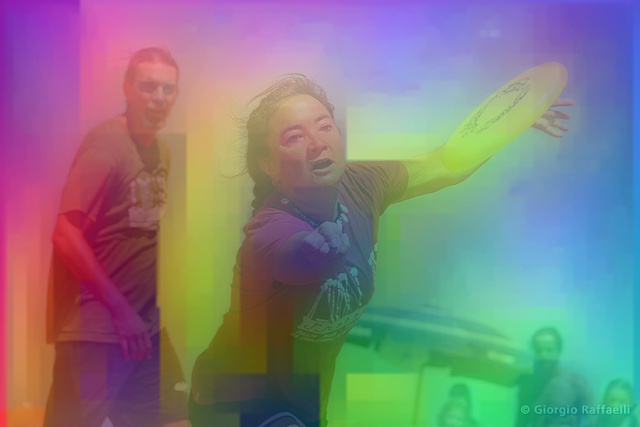}} & \raisebox{-.5\height}{\includegraphics[width=0.2\linewidth]{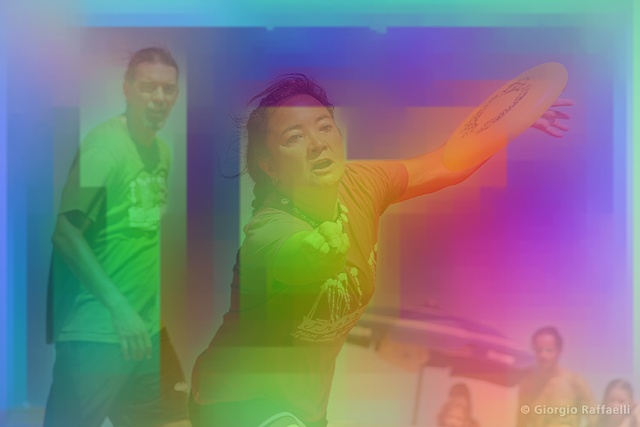}} \\ 
Sim   & \raisebox{-.5\height}{\includegraphics[width=0.2\linewidth]{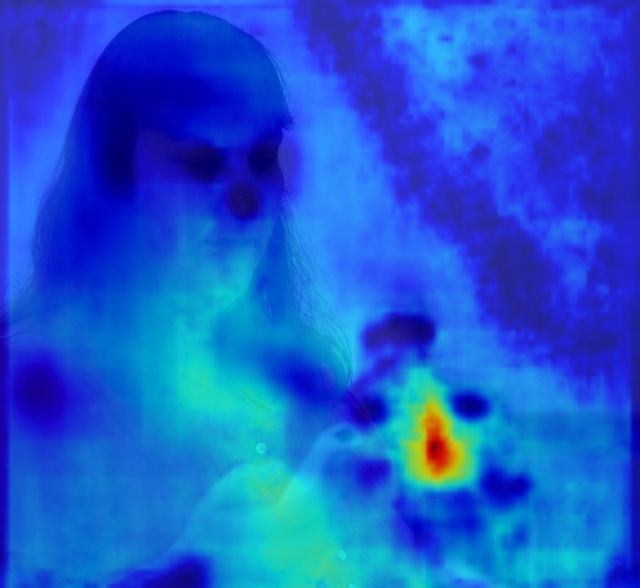}} & \raisebox{-.5\height}{\includegraphics[width=0.2\linewidth]{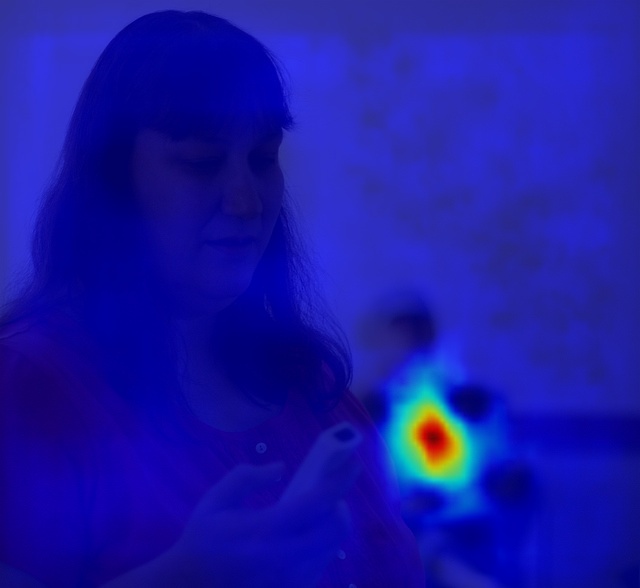}} & \raisebox{-.5\height}{\includegraphics[width=0.2\linewidth]{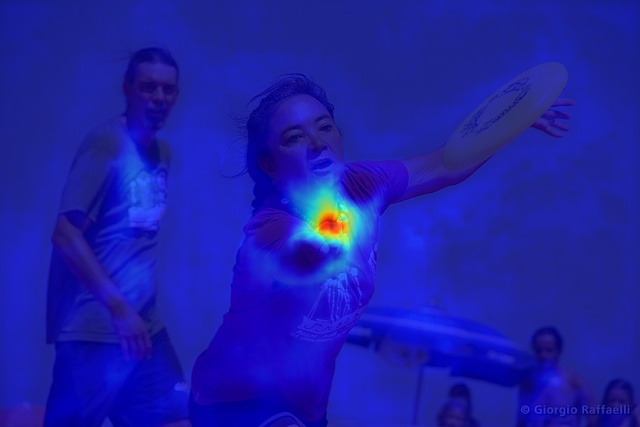}} & \raisebox{-.5\height}{\includegraphics[width=0.2\linewidth]{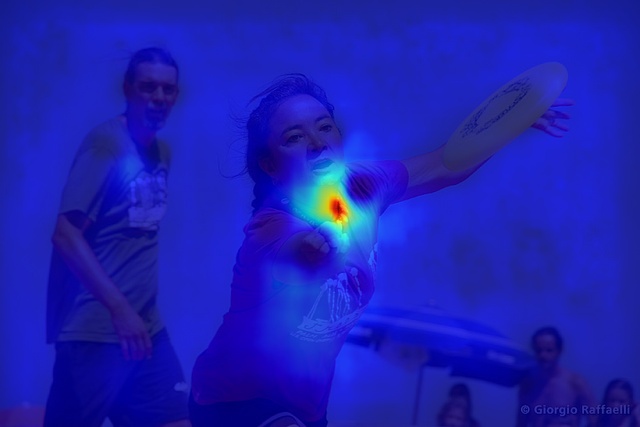}} \\ 
\end{tabular}
\end{center}
\caption{Failure cases on COCO \texttt{val} set.}
\label{fig:supp_vis_failure}
\end{figure}

We additionally report failure cases of BoIR in Fig.~\ref{fig:supp_vis_failure}. In case of the left images, BoIR produces duplicated predictions on the same person, due to wide activation area of the center heatmap. In case of the right images, BoIR places occluded joints on implausible positions, while this does not affect evaluation performance. However, BoIR generally produces disentangled instance features and detects people better than CID.

%% file: tables/coco_val_full.tex
\begin{table}[!t]
\begin{center}
\begin{tabularx}{\linewidth}{|llXXXXXXX|}
\hline
\multicolumn{1}{|l|}{Method}         & \multicolumn{1}{l|}{Backbone}   & \multicolumn{1}{c|}{Input size} & AP & AP$^{50}$ & AP$^{75}$ & AP$^{M}$ & \multicolumn{1}{c|}{AP$^{L}$} & AR \\ \hline
\multicolumn{9}{|c|}{Top-down methods}                                                                                                                            \\ \hline
\multicolumn{1}{|l|}{SBL~\cite{xiao2018simple}} & \multicolumn{1}{l|}{ResNet-152} & \multicolumn{1}{c|}{384$\times$288}    & 74.3 & 89.6 & 81.1 & 70.5 & \multicolumn{1}{l|}{81.6}    & 79.7  \\
\multicolumn{1}{|l|}{HRNet~\cite{sun2019deep}}          & \multicolumn{1}{l|}{HRNet-W32}  & \multicolumn{1}{c|}{384$\times$288}    & 75.8 & 90.6 & 82.5 & 72.0 & \multicolumn{1}{l|}{82.7}    & 80.9  \\ \hline
\multicolumn{9}{|c|}{Bottom-up methods}                                                                                                                           \\ \hline
\multicolumn{1}{|l|}{HrHRNet~\cite{Cheng_2020_CVPR}}    & \multicolumn{1}{l|}{HrHRNet-W32} & \multicolumn{1}{c|}{512}   & 67.1 & 86.2 & 73.0 & 61.5 & \multicolumn{1}{l|}{76.1}    & -  \\
\multicolumn{1}{|l|}{HrHRNet~\cite{Cheng_2020_CVPR}}    & \multicolumn{1}{l|}{HrHRNet-W48} & \multicolumn{1}{c|}{640}   & 69.9 & 87.2 & 76.1 & 65.4 & \multicolumn{1}{l|}{76.4}    & -  \\
\multicolumn{1}{|l|}{DEKR~\cite{Geng_2021_CVPR}}           & \multicolumn{1}{l|}{HRNet-W32}  & \multicolumn{1}{c|}{512} & 68.0 & 86.7 & 74.5 & 62.1 & \multicolumn{1}{l|}{77.7}    & 73.0 \\
\multicolumn{1}{|l|}{DEKR~\cite{Geng_2021_CVPR}}           & \multicolumn{1}{l|}{HRNet-W48}  & \multicolumn{1}{c|}{512} & 71.0 & 88.3 & 77.4 & 66.7 & \multicolumn{1}{l|}{78.5}    & 76.0 \\
\multicolumn{1}{|l|}{SWAHR~\cite{Luo_2021_CVPR}}          & \multicolumn{1}{l|}{HrHRNet-W32} & \multicolumn{1}{c|}{512}   & 67.1 & 86.2 & 73.0 & 61.5 & \multicolumn{1}{l|}{76.1}    & - \\
\multicolumn{1}{|l|}{SWAHR~\cite{Luo_2021_CVPR}}          & \multicolumn{1}{l|}{HrHRNet-W48} & \multicolumn{1}{c|}{640}   & 69.9 & 87.2 & 76.1 & 65.4 & \multicolumn{1}{l|}{76.4}    & - \\ \hline
\multicolumn{9}{|c|}{Single stage methods}                                                                                                                        \\ \hline
\multicolumn{1}{|l|}{PETR~\cite{Shi_2022_CVPR}}           & \multicolumn{1}{l|}{ResNet-101}     & \multicolumn{1}{c|}{800} & 70.0 & 88.5 & 77.5 & 63.6 & \multicolumn{1}{l|}{79.4}    & - \\
\multicolumn{1}{|l|}{ED-Pose~\cite{yang2023explicit}}        & \multicolumn{1}{l|}{ResNet-50}      & \multicolumn{1}{c|}{800} & 71.6 & 89.6 & 78.1 & 65.9 & \multicolumn{1}{l|}{\textbf{79.8}}    & - \\
\multicolumn{1}{|l|}{CID~\cite{Wang_2022_CVPR}}            & \multicolumn{1}{l|}{HRNet-W32}      & \multicolumn{1}{c|}{512}    & 69.8 & 88.5 & 76.6 & 64.0 & \multicolumn{1}{l|}{78.9}    & 75.4 \\ \hline
\multicolumn{1}{|l|}{BoIR}        & \multicolumn{1}{l|}{HRNet-W32}      & \multicolumn{1}{c|}{512} & \textbf{70.6} & \textbf{89.2} & \textbf{77.4} & \textbf{65.1} & \multicolumn{1}{l|}{\textbf{79.0}} & \textbf{76.3} \\
\multicolumn{1}{|l|}{BoIR}        & \multicolumn{1}{l|}{HRNet-W48}      & \multicolumn{1}{c|}{640} & \textbf{72.5} & \textbf{89.9} & \textbf{79.1} & \textbf{68.2} & \multicolumn{1}{l|}{79.4} & \textbf{78.3} \\ \hline
\end{tabularx}
\end{center}
\caption{Comparison with state-of-the-art methods on COCO \texttt{val} set. Best scores are marked as bold for small(e.g. HRNet-W32) and large(e.g. HRNet-W48) models respectively.}
\label{table:coco_val_full}
\end{table}

%% file: tables/beta_dim_ablation.tex
\begin{table}[!t]
\begin{center}
\begin{tabular}{|c|cc|}
\hline
$\beta$ & AP & AR \\ \hline
0.5 & 69.8 & 75.7 \\
1 & 70.3 & 76.2 \\
5 & 70.2 & 76.0\\
10 & 70.5 & 76.2\\
15 & 70.2 & 76.0\\ \hline
\end{tabular} \quad
\begin{tabular}{|c|cc|}
\hline
Emb. Dim & AP & AR \\ \hline
1 & 70.4 & 76.3 \\
8 & 70.4 & 76.3 \\
16 & 70.4 & 76.2 \\
32 & 70.4 & 76.3\\
64 & 70.1 & 75.8 \\
128 & 70.5 & 76.2 \\ \hline
\end{tabular}
\end{center}
\caption{Left: Ablation experiment of $\beta$ on COCO \texttt{val} set. $D=128$ by default. Right: Ablation experiment of embedding dimension $D$ on COCO \texttt{val} set. $\beta=10$ by default.}
\label{table:ablation_beta_dim}
\end{table}